\documentclass{article}

\usepackage[main, final, nonatbib, preprint]{neurips_2026}

\usepackage{caption}
\usepackage{wrapfig}
\usepackage{amsmath}
\usepackage{amssymb}
\usepackage{enumitem}
\usepackage{threeparttable}
\usepackage{multirow}
\usepackage{algorithm}
\usepackage{algorithmic}
\usepackage{subfigure}
\usepackage[many]{tcolorbox}
\usepackage{booktabs}
\usepackage{color}

\usepackage{hyperref}
\hypersetup{
  colorlinks=true,
  linkcolor=blue,
  filecolor=magenta,
  urlcolor=cyan,
}

\newcommand{\appoutlineitem}[2]{%
  \item \hyperref[#1]{\ref*{#1}\enspace #2}\dotfill\pageref{#1}%
}%
\newcommand{\ourname}{{ReMem}} 
\newcommand{\improve}{{7.58}} 
\newcommand{\featno}{\textcolor{red}{\texttimes}}
\newcommand{\featyes}{\textcolor{green!60!black}{\checkmark}}

\usepackage{tabularx}
\newcolumntype{C}{>{\centering\arraybackslash}X}

\usepackage[utf8]{inputenc}%
\usepackage[T1]{fontenc}%
\usepackage{hyperref}%
\usepackage{url}%
\usepackage{booktabs}%
\usepackage{amsfonts}%
\usepackage{nicefrac}%
\usepackage{microtype}%
\usepackage{xcolor}%

\title{ReMem: Rethinking Perception and Memory in Long-Context Recommendation Agents}

\author{%
  Haohao~Qu\thanks{This work is completed during Haohao Qu’s research attachment at NTU.} \\
  The Hong Kong Polytechnic University\\
  Nanyang Technological University\\
  \texttt{haohao.qu@connect.polyu.hk} \\
  \And
  Yongcheng Jing\thanks{Corresponding authors.} \\
  Nanyang Technological University \\
  \texttt{yongcheng.jing@ntu.edu.sg} \\
  \AND
  Chun Hin CHAN \\
  The Hong Kong Polytechnic University\\
  \texttt{chun-hin-vincent.chan@connect.polyu.hk} \\
  \And
  Shanru Lin \\
  The Hong Kong Polytechnic University\\
  \texttt{lllam32316@gmail.com} \\
  \And
  Wenqi Fan\footnotemark[2] \\
  The Hong Kong Polytechnic University\\
  \texttt{wenqifan03@gmail.com} \\
  \And
  Dacheng Tao\footnotemark[2] \\
  Nanyang Technological University \\
  \texttt{dacheng.tao@ntu.edu.sg} \\
}

\begin{document}

\maketitle

\begin{abstract}
Recent Recommendation Agents (RecAgents) offer a promising alternative by shifting recommendation to an active, user-side paradigm, where generative agents autonomously perceive web platforms, reason over user preferences, and execute decisions.
However, existing RecAgents still suffer from two critical limitations: brittle item perception based on noisy item pages, and inefficient long-context reasoning over extended user histories and multi-step interaction traces.
To address these challenges, we propose a recommendation-agent framework, termed \textbf{\ourname{}}, that combines OCR-based multimodal perception with time-evolving dynamic memory.
Instead of parsing raw HTML, \ourname{} observes item-page screenshots and uses an OCR tool to produce structured textual evidence for the reasoning backbone.
To support long-horizon preference modeling, \ourname{} further introduces a chunk-wise sequential update strategy that maintains a bounded textual memory while processing histories longer than the backbone context window; with fixed chunk and memory budgets, the number of update calls grows linearly with history length.
This design allows the agent to preserve evolving user preferences without relying on external memory modules or disrupting the standard autoregressive generation process.
To train the dynamic memory updater, we further develop a multi-memory GRPO variant whose reward combines recommendation-specific information with final-task metrics.
Experiments on three Amazon-derived datasets and a controlled post-retrieval protocol show an average relative improvement of \improve\% over the strongest reported baseline across searching, ranking, and judging tasks. 
\url{https://github.com/Quhaoh233/ReMem}
\end{abstract}

\section{Introduction}

\begin{figure}
    \centering
    \includegraphics[width=0.75\linewidth]{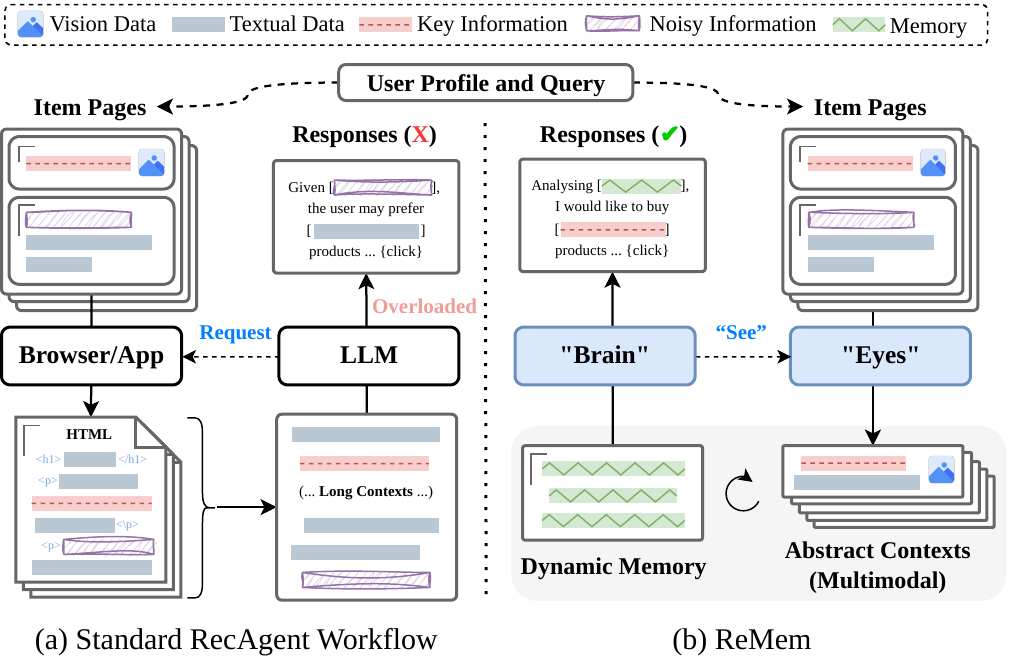}
      \captionsetup{font={small}}
\caption{
Motivation of \ourname{}.
(a) Conventional RecAgents read the Web as text: they extract noisy HTML into long contexts and rely on LLMs to recover user intent, which is brittle across modalities and inefficient for long-context reasoning.
(b) The proposed \ourname{} follows a more human-like principle: \emph{see} item pages through OCR-based multimodal perception and \emph{memorize} only evolving preference abstractions through dynamic memory, enabling robust modeling of time-evolving user preferences.
}
    \label{fig:intro}
\end{figure}

The rapid advancement of Large Language Models (LLMs) has sparked a surge of interest in LLM-based Recommender Systems (RSs)~\cite{geng2022recommendation,rajput2023recommender,bao2023tallrec,liao2024llara,wang2024rethinking}, driven by their exceptional generalization capabilities and remarkable proficiency in in-context learning~\cite{zhao2024recommender}.
These unique properties empower LLM-based recommenders to generalize effectively across unseen tasks and diverse domains, thereby significantly enhancing overall recommendation quality and system utility~\cite{wang2025knowledge}.
However, despite opening a promising frontier for personalized recommendation applications, deploying LLMs in traditional recommender architectures is severely bottlenecked by massive token consumption and high latency under high-concurrency traffic~\cite{hua2023index,qu2025tokenrec,hou2025generating}.
These computational bottlenecks notably impair the practical deployment of LLM-based systems in standard {passive} recommendation scenarios, where a platform-side model must deliver real-time, low-latency personalized recommendations to millions of passive users~\cite{kang2018self,fan2019graph,zhang2025ssd4rec}.
To circumvent these deployment constraints and leverage the cognitive capabilities of LLMs more naturally, researchers have turned to the paradigm of autonomous AI agents.
Inspired by the recent success of LLM-based agents such as Anthropic\footnote{\url{https://www.anthropic.com}} and OpenClaw\footnote{\url{https://openclaw.ai}}, emerging research on personal Recommendation Agents (\emph{RecAgents}) has gained rapid traction~\cite{xu2025iagent,huang2025recommender,zhang2024agentcf,xia2026multi,shang2026agentrecbench}.
Instead of forcing computationally heavy LLMs to handle real-time, platform-side passive recommendations, the agent paradigm shifts the application to a more suitable {active} \textbf{Web Mining} setting on the user side.
Here, users express open-ended goals, and personalized AI agents autonomously handle the ``last mile'' of the decision-making process by interacting with external web platforms.
As illustrated in Figure~\ref{fig:intro} (a), the workflow of a typical RecAgent aligns with general agent architectures~\cite{sumers2024cognitive} and can be categorized into four core modules: 
(i) \emph{perception}, which senses user intent and extracts candidate items; 
(ii) \emph{reasoning and planning}, which models user preferences based on historical interactions and queries; 
(iii) \emph{action}, which interacts with target websites and  executes recommendation decisions.

Despite their potential, existing RecAgents face two fundamental limitations when deployed in complex, real-world recommendation environments.
\textbf{First, how should a RecAgent perceive items?}
Existing approaches predominantly rely on reading item web pages in raw HTML format~\cite{deng2023mind2web,yao2022webshop,gur2024real,chae2025web}. 
This paradigm often fails when agents navigate across noisy item pages, misses rich multimodal details beyond plain text, and introduces significant noise (e.g., advertisements and tracking scripts) that degrades generation quality.
\textbf{Second, how should a RecAgent handle long-context reasoning?}
Retrieving multi-step actions, executing complex reasoning chains, and managing long-term user interaction histories generate an overwhelming volume of tokens that easily exceeds the finite context windows of current LLMs~\cite{shi2025personax,liu2025agentcf++,wang2026vtc}.
While current LLM-based RSs seek to condense information via token-level reduction~\cite{qu2025tokenrec,qu2026diffusion} or external memory plugins~\cite{wang2025knowledge,huang2025recommender,zhang2024generative}, they often struggle with out-of-distribution generalization. Furthermore, they require auxiliary modules or complex context operations that inevitably disrupt the standard autoregressive generation process, thereby hindering compatibility and parallelization.

These limitations prompt a fundamental rethinking of the paradigm underlying RecAgents:
\emph{Can a recommender agent uniformly capture multimodal item information amid noisy distractions, while maintaining resource-efficient, long-context reasoning over extended user histories?}
To address this, we draw inspiration from two anthropocentric intuitions.
First, regarding \emph{perception}, web designers deliberately optimize user interfaces to be visually intuitive for humans, highlighting key product specifications while pushing advertisements to the periphery.
Humans naturally scan these pages visually rather than parsing raw HTML code.
Thus, a screenshot-based perception mechanism can reduce dependence on the HTML schemas.
Second, regarding \emph{memory and reasoning}, when humans process long-term information, we do not attempt to memorize every trivial detail.
Instead, we dynamically abstract key concepts, update our cognitive state, and discard redundant information to manage cognitive load.
This selective attention allows us to solve complex, long-horizon tasks with bounded memory capacity.

Motivated by these intuitions, we introduce a novel RecAgent framework, named \textbf{\ourname{}}, that leverages {Optical Character Recognition (OCR)-based multimodal perception} for humanoid information gathering, coupled with dynamic memory enhanced by Reinforcement Learning (RL) for long-context reasoning.
Specifically, we integrate DeepSeek-OCR-2~\cite{wei2025deepseek,wei2026deepseek} as an active tool-use component to achieve human-like multimodal perception.
By parsing screenshots of item pages, this approach extracts structured multimodal item information, effectively bypassing the noisy and brittle nature of HTML parsing.
Furthermore, we propose a chunk-wise, dynamic memory update mechanism that aligns with evolving user preferences over time.
During reasoning, the backbone LLM processes the OCR text sequentially in chunks.
As it reads each chunk, the model proactively and selectively updates a fixed-length memory containing $K$ historical interactions.
Finally, to teach the agent to accommodate this multi-turn memory mechanism, we develop a simple but effective variant of GRPO~\cite{shao2024deepseekmath,yu2026dapo} that incorporates recommendation-specific information and evaluation metrics into its reward function.

In summary, our main contributions are as follows:
\begin{itemize}[leftmargin=*]
\item This study revisits two fundamental questions in long-context recommendation agents: \emph{How should a RecAgent ``see'' and ``memorize''?}
We find that the raw HTML content and static memory used by most existing studies are insufficient to capture multimodal item information and model users' time-evolving preferences over long interaction histories.

\item We propose a novel RecAgent framework termed \ourname{}.
To the best of our knowledge, this is the first framework to integrate an OCR-based perception module with RL-enhanced recurrent memory for time-evolving preference modeling in long-context RecAgents; the novelty lies in this recommendation-specific integration rather than in any individual component.

\item Experiments on three Amazon-derived \emph{InstructRec}~\cite{zhang2026recommendation} datasets evaluate \emph{Searching}, \emph{Ranking}, and \emph{Judging} under a controlled 10-candidate protocol. Across the 12 reported task--dataset metrics, \ourname{} achieves an average relative improvement of \improve\% over the strongest reported baseline.
\end{itemize}

\section{Pilot Study}
\label{sec:pilot}
In this section, we systematically examine the challenges in perception and long-context reasoning. By isolating these capabilities from other components of the recommendation pipeline, the controlled pilot studies enable more precise diagnosis and reduce confounding factors that would otherwise obscure the sources of performance degradation. The findings motivate the design of the proposed \ourname{} framework in Section~\ref{sec:method}.

\subsection{Preliminary}
A standard pipeline for applying AI agents to recommender systems typically consists of several key modules~\cite{wang2024recmind}: an LLM backbone, such as Qwen, that drives the overall reasoning process; perception tools that obtain external information from web pages when such information is absent from the model parameters; planning and reasoning modules that decompose a task into smaller sub-tasks and solve them step by step; and action modules that interact with the Internet, e.g., through search and clicks, to produce the final recommendations.
We consider the setting where the goal is to generate a final recommendation result $y$ given several key inputs, such as a user query $\mathcal{Q}$, interaction history $\mathcal{H}$, and candidate items $\mathcal{Z}$, using an LLM-based agent parameterized by $\theta$. The standard workflow can be formulated as
\begin{align}
\label{eq:preliminary}
    y \sim p_\theta \bigl(y \mid \texttt{reasoning}(\texttt{planning}(\mathcal{Q}), \mathcal{H}), \texttt{perception}(\mathcal{Z})\bigr),
\end{align}
where $\texttt{planning}(\mathcal{Q})$ denotes a set of prompts that decomposes the problem $\mathcal{Q}$ into a series of sub-tasks, $\texttt{reasoning}(\mathcal{H})$ is the user modeling process through LLMs based on interaction history, and $\texttt{perception}(\mathcal{Z})$ represents the search and content-recognition tools used to collect information about potential items.

In the following subsections, we analyze the impact of agent perception and reasoning mechanisms on \textbf{\emph{WebWalkerQA}}\footnote{\url{https://huggingface.co/datasets/callanwu/WebWalkerQA}}~\cite{wu2025webwalker} and \textbf{\emph{HotpotQA}}\footnote{\url{https://huggingface.co/datasets/BytedTsinghua-SIA/hotpotqa}}~\cite{yang2018hotpotqa}, respectively.
WebWalkerQA is a general benchmark for evaluating LLMs in web traversal.
It contains 673 QA examples that require a model to access the Internet and retrieve information from websites containing both visual and textual content.
HotpotQA evaluates a model's ability to locate and extract relevant information from realistic document collections, with context lengths of approximately [7K - 3.5M] tokens.
Detailed experimental configurations are provided in Appendix~\ref{app:ocr_eval}.

\subsection{Revisiting Agent Perception}

This subsection investigates how different perception strategies affect a standard LLM-agent workflow on the WebWalkerQA dataset, which contains real webpages with diverse layouts.

\noindent\textbf{Experimental Setting}.
A typical LLM reasoning process feeds plain textual information, i.e., raw content, into the LLM backbone under an in-context learning paradigm.
Based on this setting, we compare three perception strategies:
1) performing actions and retrieving information from real websites by converting the ``mainbody'' of HTML snippets, together with website images recorded as URL links, into readable prompt templates~\cite{gur2024real,koh2024visualwebarena};
2) taking a screenshot of the page as image input for the agent to respond to the user query~\cite{zhang2026vipact};
and 3) parsing the text and images on the website or app page using DeepSeek-OCR-2~\cite{wei2026deepseek}, and then embedding the OCR results into prompt templates.
For each strategy, we repeat the evaluation five times to reduce random variation.

\begin{wraptable}{r}{0.52\linewidth}\scriptsize
\centering
\captionsetup{font={small}}
    \caption{The effect of different perception strategies.
    Strategy 3) parses multimodal web pages into informative textual descriptions through OCR perception, achieving superior performance compared with the other strategies.}
    \begin{tabular}{ccccccc}
        \toprule
       \multirow{2}{*}{Acc. (\%)} & \multicolumn{3}{c}{{Qwen3.5}} & \multicolumn{3}{c}{{Qwen3VL}} \\ \cmidrule(lr){2-4} \cmidrule(lr){5-7}
        & {4B}      & {9B}      & {Avg.}      & {4B}      & {8B}      & {Avg.}      \\ \midrule
        Baseline &  49.13 & 54.27 & 51.70 & 49.32 & 50.94 & 50.13  \\ \midrule
        \quad + \textit{1)} & 41.51 & 45.27 & 43.39 & 52.51 & 55.77 & 54.14  \\
        \quad + \textit{2)} &  -  &    -  &   -  & 46.38 & 48.56 & 47.47   \\
       \quad \textbf{+ \textit{3)}} & \textbf{57.02} & \textbf{60.25} & \textbf{58.64} & \textbf{56.62} & \textbf{58.60} & \textbf{57.61}    \\
        \bottomrule
    \end{tabular}
    \label{tab:pilot_perception}
\end{wraptable}

\noindent\textbf{Observations}.
As shown in Table~\ref{tab:pilot_perception}, parsing websites and retrieving key web information through OCR significantly improves performance on both the WebWalkerQA dataset.
Compared with directly using the full HTML context or a website screenshot, OCR serves as the ``eyes'' of the LLM backbone: it captures rich multimodal information while filtering noisy content, such as advertisements. This allows the model to focus more on reasoning rather than perception.
Nevertheless, OCR-based perception also introduces a challenge for the long-context capabilities of LLMs~\cite{liu2024lost}. Performance may degrade when the model must retrieve relevant information from the middle of a long OCR-derived context. In other words, seeing more is not necessarily useful if the model cannot effectively reason over what it sees. This problem becomes more pronounced in recommendation scenarios, where LLMs must process long OCR perception contexts from multiple candidate items together with historical user interactions.

\noindent\textbf{Insights}.
OCR helps the agent perception module acquire richer information from multimodal web pages, but it also places greater demands on the LLM's long-context reasoning ability.

\subsection{Rethinking Long-Context Reasoning}
Long-context reasoning~\cite{liu2024lost,hsieh2024ruler} is central to agent and RecAgent workflows, since these systems often need to retrieve information across multiple steps, execute complex reasoning chains, and manage long-term user interaction histories~\cite{yu2026memagent,chhikara2025mem0}.
In this subsection, we investigate the impact of different reasoning mechanisms across varying context lengths on HotpotQA under a ``Needle-in-a-Haystack'' setting.

\noindent\textbf{Experimental Setting}.
A straightforward and widely used context-management strategy in LLM agents is to append all previous information, such as observations, intermediate thoughts, and actions, to the prompt at each interaction turn~\cite{yao2023react,zhou2026mem}. Based on this baseline, we analyze three strategies for long-context management:
1) using a specially designed long-context LLM, QwenLong-L1-32B\footnote{\url{https://huggingface.co/Tongyi-Zhiwen/QwenLong-L1-32B}}~\cite{wan2025qwenlong};
2) vectorizing documents into an external vector database and retrieving the top-5 document chunks, each containing 5K tokens, according to their semantic search scores with respect to the query, using a semantic search model\footnote{\url{https://huggingface.co/sentence-transformers/multi-qa-mpnet-base-cos-v1}}~\cite{xu2026mem,zhang2025memevolve};
and 3) using dynamic memory~\cite{yu2026memagent}, a human-like memory mechanism in which the LLM itself captures and abstracts key information from each document chunk into a shared fix-length memory cache.
We also include a truncation baseline that randomly samples contexts and truncates them to a fixed length of 128K tokens.

\begin{wrapfigure}{r}{0.6\linewidth}\scriptsize
    \centering
    \includegraphics[width=\linewidth]{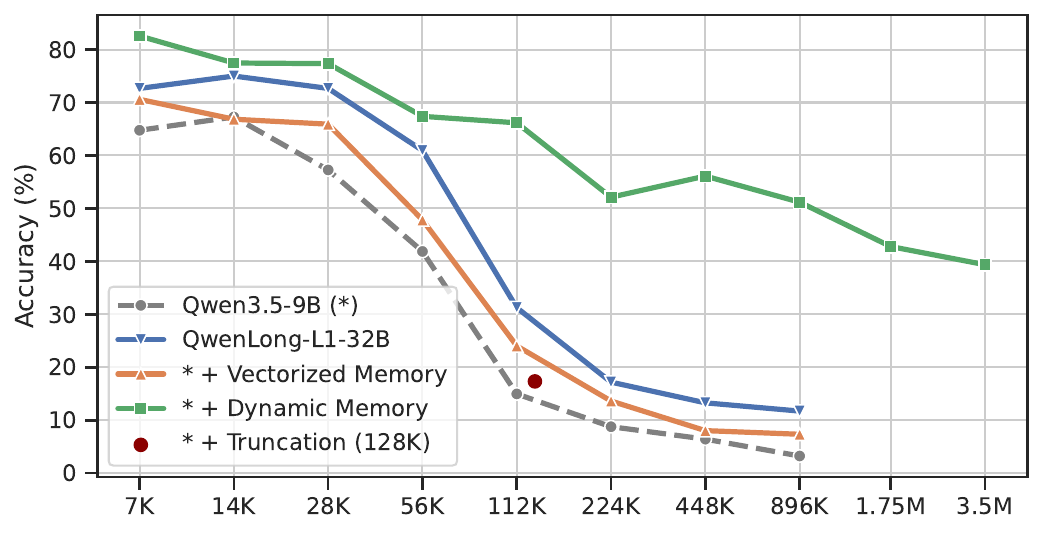}
      \captionsetup{font={small}}
    \caption{Accuracy scores of different long-context reasoning strategies on HotpotQA. Even models that use long-context continual pretraining or vectorized memory techniques fail to maintain consistent performance. In contrast, the agent equipped with dynamic memory demonstrates relative lossless performance extrapolation.}
    \label{fig:pilot}
\end{wrapfigure}

\noindent\textbf{Observations}.
Figure~\ref{fig:pilot} shows the relationship between long-context reasoning mechanisms and performance. The baseline model exhibits rapid performance degradation. QwenLong-L1 maintains reasonable performance within its training length of 60K tokens, but its performance drops substantially beyond this range. The model equipped with vectorized memory maintains acceptable performance up to 112K tokens. However, its accuracy deteriorates to 10\% at 896K tokens, as semantic memory retrieval becomes increasingly difficult when the memory base contains a large number of vectorized documents~\cite{zhao2026amabench}.
In contrast, dynamic memory, implemented entirely through the LLM itself, is simple yet effective. It shows strong length-extrapolation ability, with only marginal performance decay as the input context length increases in the locating-and-extracting task.

\noindent\textbf{Insights}.
The effectiveness of dynamic memory suggests its potential for agentic recommendation scenarios, where the system must identify time-evolving user preferences from user queries and manage long-term interaction histories.

\section{Methodology}
\label{sec:method}
\subsection{The Overview of the Proposed Approach}
\begin{figure*}
    \centering
    \includegraphics[width=\linewidth]{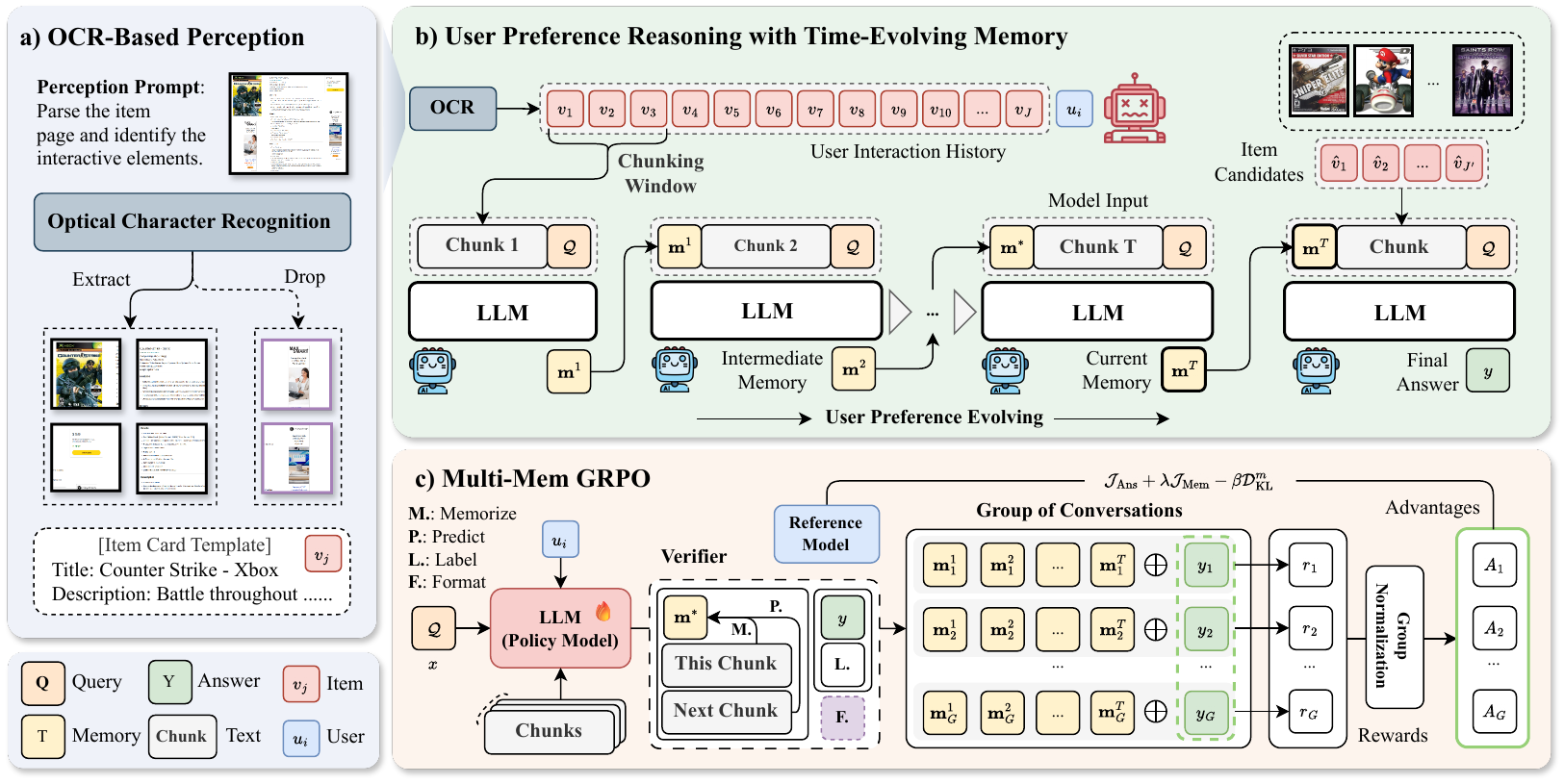}
      \captionsetup{font=small}
\caption{Overview of the proposed \ourname{} framework. Inspired by how humans browse item pages while maintaining time-evolving memory, \ourname{} consists of three key components. a) Instead of relying on raw HTML, \ourname{} adopts an OCR-based perception module that represents item pages through a unified multimodal view of natural text, images, and interactive elements, better matching users' browsing behavior. b) User interaction history is processed as a sequential stream of chunks, from which the model maintains a compact, fixed-length memory that is continuously updated to capture evolving user preferences. c) To teach the model what to remember and how to update memory in multi-turn interactions, \ourname{} employs Multi-Memory GRPO, which incorporates recommendation-specific information for intermediate memory states and evaluation metrics for the final answer into its reward function.
}
    \label{fig:overview}
\end{figure*}

Driven by the aforementioned insights in Sect.~\ref{sec:pilot}, we propose a novel framework, \ourname{}, for recommendation AI agents.
As shown in Figure~\ref{fig:overview}, the design principle of \ourname{} lies on two anthropocentric intuitions.
First, the paper embraces a multimodal perception solution through OCR~\cite{wei2025deepseek,wei2026deepseek} to replace the original reading format (e.g., HTML), as the majority of item pages are with a human-centered design philosophy: emphasizing the key information visually.
The OCR technique can preserve essential information while filtering noisy contents (e.g., ads) to improve multimodal perception of item pages.
Second, we introduce a time-evolving memory mechanism designed for capturing the evolving user preferences.
This mechanism enables RecAgents to process inputs of any length within a finite context window in linear time complexity during the inference process, thereby overcoming a major bottleneck in long-context processing.
To better implement this mechanism, we process a multi-memory reinforcement learning (RL) algorithm, based on novel GRPO techniques~\cite{shao2024deepseekmath,yu2026memagent}.
The key modules are presented in the subsequent sections.

\subsection{Multimodal Perception on Item Pages}
Rather than relying on the raw HTML format, this paper proposes embracing a more general solution for item perception: \emph{"see" item pages with natural text, images and interactive elements in a unified perspective}. 
Recently, Optical Character Recognition (OCR) has emerged as a foundational technology capable of converting images and scanned text documents into structured templates that can be read by large language models (LLMs)~\cite{cui2025paddleocr,wei2025deepseek}.
To enable multimodal perception in agent-based recommendation scenarios, OCR is more promising than ever—with massive amounts of unstructured data being generated and consumed every day.

As shown in Figure~\ref{fig:overview} (a), after request to a certain item page, the proposed \ourname{} prompts an advanced OCR model, i.e., DeepSeek-OCR-2~\cite{wei2026deepseek}, to parse the whole page including not only the textual contents but also the images or charts
\begin{align}
    v_j = \text{OCR}(\text{Request}(\text{url}(\text{texts, images, charts}))),
\end{align}
which enriched with insights that web designers or item producers wanna users to see.
The OCR module consists of an encoder and a decoder.
The encoder discretizes item pages into visual tokens,
while the decoder generates outputs conditioned on these visual tokens and text prompts.
Finally, the OCR outputs are embedded into a template of item cards in natural language, so that they can be easily used by the LLM backbone for subsequent reasoning.
OCR errors may arise from low-quality images, complex layouts, or uncommon fonts. However, our item pages are well structured, mitigating such errors. See Appendix~\ref{app:ocr_eval} for further discussion.

\subsection{Time-Evolving Memory for User Preference}
\label{sec:tem}
RecAgents must reason over long and continuously evolving user histories and OCR-extracted item descriptions. Directly feeding the full history into an LLM is infeasible due to finite context windows and quadratic attention cost.
We therefore propose a \emph{time-evolving memory (TEM)} mechanism with three goals: supporting long histories, avoiding long-context performance degradation, and enabling linear-time inference.
The central idea is to process the interaction history as a sequential stream and maintain a compact, fixed-length memory that is updated over time.
Instead of storing every historical token, the \ourname{} selectively abstracts preference-relevant information, such as stable interests, recent intent shifts, price sensitivity, brand preference, and task-specific constraints.
This design enables the agent to reason over long-term user behavior while keeping the active context size bounded.

\subsubsection{\textbf{Workflow}}
As illustrated in Figure~\ref{fig:overview}(b), the proposed memory mechanism treats a long user interaction history as a controlled stream of evidence rather than a monolithic context.
At each step, the RecAgent observes two components: 
(1) the next chunk of historical interactions, and 
(2) a compact memory summarizing the preference evidence accumulated so far. 
The memory is represented as ordinary natural-language tokens inside the LLM context window.
Therefore, the backbone LLM does not require any architectural modification, external key-value cache manipulation, or customized attention operator.

The working procedure naturally consists of two stages: namely \emph{context-processing} and \emph{answer-generation}.
In the first stage, the model sequentially reads chunks of the user history and updates the memory after each chunk.
For a user $u_i$, the interaction history $\mathcal{H}=\{v_1,\ldots,v_J\}$ is partitioned into $T$ contiguous chunks $\mathcal{C} = \{\mathbf{c}^1,\ldots,\mathbf{c}^T \}$ based on a chunking window of $W$, where each chunk $\mathbf{c}^t=\text{join}[v_{ t W}, \dots, v_{(t+1)W}]$ may contain $W$ user--item interactions and the corresponding item descriptions.
Given the previous memory $\mathbf{m}^{t-1}$ and the current chunk $\mathbf{c}^t$, the RecAgent produces an updated memory $\mathbf{m}^{t}$:
\begin{align}
\label{eq:memory_update}
    \mathbf{m}^{t} \sim p_\theta\left(\mathbf{m}^{t} | \mathbf{m}^{t-1}, \mathbf{c}^{t}, \mathcal{Q}\right),
\end{align}
where $\mathcal{Q}$ denotes the user query or recommendation task, the memory is set to a fix-length of $|\mathbf{m}^t|=L$, and the initial memory $\mathbf{m}^0=\emptyset$.
The updated memory overwrites the previous one.

In the \emph{answer-generation} stage, after all chunks have been processed, \ourname{} generates the final recommendation by conditioning only on the task input, candidate information, and the final memory:
\begin{align}
\label{eq:final_output}
    y \sim p_\theta\left(y | \hat{\mathbf{c}}, \mathbf{m}^{T}, \mathcal{Q}\right),
\end{align}
where $\hat{\mathbf{c}} = \texttt{perception}(\mathcal{Z})$ denotes the OCR-based perception results of potential candidate items $\mathcal{Z}=\{\hat{v}_1,\dots, \hat{v}_{J'}\}$.

This design has three advantages. First, it supports \emph{unbounded histories}: the interaction sequence can be long because it is processed chunk by chunk. Second, it mitigates the \emph{long-context performance cliff}: instead of forcing the model to attend to all past tokens, the memory keeps only preference-relevant evidence. Third, it ensures \emph{linear cost}: because the chunk size and memory size are fixed, the computational cost grows linearly with the number of chunks. As a result, a moderately context-sized LLM can be converted into an efficient long-context preference reasoner with minimal engineering overhead.

\subsubsection{\textbf{Derivation}}
A standard autoregressive LLM~\cite{zhao2023survey} factorizes the likelihood of a token sequence $\mathbf{x}_{1:N}$ as
\begin{align}
\label{eq:auto-regressive}
    p(\mathbf{x}_{1:N})
    =
    \prod_{n=1}^{N} p(x_n | \mathbf{x}_{1:n-1}).
\end{align}
This formulation implicitly assumes that all previous tokens, or their hidden states, are available in the active context. When $\mathbf{x}_{1:N}$ corresponds to a long user history, this assumption becomes impractical because the attention cost grows quadratically with the context length $N$.

Equivalently, the long-history recommendation process can be viewed as marginalizing over a sequence of latent memory states $\mathbf{m}^{1:T}$, which decomposes the original likelihood in Equation~\eqref{eq:auto-regressive} as
\begin{align}
&p_{\theta}\!\left(
y\mid\mathcal{Q},\mathbf{c}^{1:T},\hat{\mathbf{c}}
\right) \nonumber\\
&=
\sum_{\mathbf{m}^{1:T}}
\left[
\prod_{t=1}^{T}
p_{\theta}\!\left(
\mathbf{m}^{t}
\mid
\mathbf{m}^{t-1},\mathbf{c}^{t},\mathcal{Q}
\right)
\right]
p_{\theta}\!\left(
y\mid\mathbf{m}^{T},\hat{\mathbf{c}},\mathcal{Q}
\right).
\end{align}
Inside each chunk, we still run an ordinary transformer decoder, but conditioned on a constant context window ($\mathbf{c}^t$, $\mathbf{m}^{t-1}$).
In which, the memory update path in Equation~\eqref{eq:memory_update} factorizes token-by-token via the backbone LLM $\theta$:
\begin{align}
\label{eq:tem}
    p_\theta\left(\mathbf{m}^{t}\mid \mathbf{m}^{t-1},\mathbf{c}^{t},\mathcal{Q}\right)
    =
    \prod_{l=1}^{L}
    p_\theta\left(m^{t}_{l}
    \mid
    \mathbf{m}^{t}_{<l},
    \mathbf{m}^{t-1},
    \mathbf{c}^{t},
    \mathcal{Q}
    \right).
\end{align}
This equation shows that the TEM mechanism does not need to condition on the entire history explicitly.
Instead, it propagates user preference through a sequence of compact memory states.
Conceptually, this transforms the transformer decoder into a recurrent preference updater whose state size is controlled by the memory budget $M$.
Moreover, no special positional embedding extrapolation, attention re-scaling, or non-standard cache operation is required.

\subsubsection{\textbf{Discussions}}
Compared with feature-space compression methods such as linear attention or hidden-state summarization, our memory is explicitly represented in token space. This property is particularly useful for recommendation agents: the intermediate memory can be inspected, debugged, edited, or constrained by prompts. For example, the memory may explicitly record that a user prefers lightweight running shoes, dislikes overly expensive items, recently searched for waterproof products, and tends to choose neutral colors. Such human-readable states improve transparency and make the agent's recommendation process easier to control.
In summary, the time-evolving memory provides a practical mechanism for long-horizon user preference modeling. It allows the RecAgent to process long histories, maintain a bounded context window, and preserve the standard autoregressive generation procedure of the backbone LLM.

\subsection{Multi-Memory GRPO}
\label{sec:multi_conv_grpo}

In our setting, a single query induces multiple conversations, which are used as intermediate memories for the final reasoning process, as illustrated in Figure~\ref{fig:overview}.
This differs from standard multi-turn tool-calling optimization~\cite{yu2026memagent}, where the entire trajectory can be treated as one sequential context and optimized with an attention mask.
Since our conversations are independently generated and later aggregated, directly applying the standard token mask would not correctly assign credit to the intermediate memories.
To address this, we propose \emph{Multi-Memory GRPO}, a variant of GRPO~\cite{shao2024deepseekmath} that incorporates recommendation-specific information and evaluation metrics into its reward function.

\subsubsection{\textbf{Multi-Memory Generation}}
Let $\mathcal{Q}$ denote an input query, $\hat{\mathbf{c}}$ is the chunk for candidates $\mathcal{Z}$, and $\mathcal{C}=\{\mathbf{c}^1, \mathbf{c}^2, \dots, \mathbf{c}^T\}$ represents the chunked contexts of interacted items for the user $u_i$.
As illustrated in Section~\ref{sec:tem} and Equation~\eqref{eq:memory_update},
for each query, a policy model $\pi_\theta$ generates $T$  sequential memory instances:
\begin{equation}
    \mathcal{M}
    =
    \{\mathbf{m}^1, \mathbf{m}^2, \dots, \mathbf{m}^T\},
    \quad
    \mathbf{m}^{t} \sim \pi_{\theta_{\mathrm{old}}}(\cdot \mid \mathbf{m}^{t-1}, \mathbf{c}^t, \mathcal{Q}),
\end{equation}
where the base memory $\mathbf{m}^0=\emptyset$.
These memory summaries are then used by the model to produce a final answer: $y \sim \pi_{\theta_{\mathrm{old}}}(\cdot \mid \mathbf{m}^T, \hat{\mathbf{c}}, \mathcal{Q})$ using  the accumulated memory $\mathbf{m}^T$.

Building upon this framework, we can sample a group of $G$ complete reasoning instances with a high generation temperature (e.g., 1.5) to make model outputs be more creative:
\begin{equation}
    \{\mathcal{M}_g, y_g\}^G_{g=1}, \quad \mathcal{M}_g = \{\mathbf{m}^1_g, \dots,\mathbf{m}^T_g\},
\end{equation}
where $\mathcal{M}_g$ denotes the set of intermediate memory for the $g$-th sampled reasoning instance. For every non-terminal update $t<T$, we require the memory to contain two explicit fields,
$
    \mathbf{m}_g^t
    =\big[\mathbf{s}_g^t;\,\hat{\mathcal{K}}_g^{t+1}\big] $,
where $\mathbf{s}_g^t$ summarizes the preference evidence accumulated through previous chunks, and $\hat{\mathcal{K}}_g^{t+1}$ predicts the user's preferences. Here, $\mathcal{K}^{t+1}$ denotes the set of keywords extracted from the item features (e.g., category, type, and brand) in the next chunk.

\subsubsection{\textbf{Reward Function and Group Advantage}}
To make sure the agent ground into our recommendation scenarios, each sampled instance $g \in G$ is evaluated using recommendation information and metrics beyond typical formatting rewards $r_f$.
Specifically, for intermediate memory, we use set $F1$ score to reward both correct keywords and adequeate coverage: $r^{m}_{g,t} = F1{(\mathcal{K}^{t+1},\hat{\mathcal{K}}^{t+1}_g)} + r^m_f$, which evaluates whether the memory contains the relevant recommendation information. The final answer reward is defined as $r^a_g = \mathbf{1}_\text{is\_equiv}(y^+,y_g) + r^a_f$, where $y^+$ is the title of ground-truth item.
For more details, see Appendix~\ref{app:reward}.
For any reward $\{r_g\}^G_{g=1}$, we compute the group-relative advantage by normalizing rewards across the $G$ sampled instances:
\begin{equation}
    \text{GN}(r_g) =
    \frac{r_g - \mu_r}{\sigma_r + \epsilon},
    \
    \mu_r = \frac{1}{G}\sum_{g=1}^{G} r_g, \
    \sigma_r =
    \sqrt{
    \frac{1}{G}
    \sum_{g=1}^{G}
    (r_g - \mu_r)^2
    },
\end{equation}
where $\epsilon$ is a small constant for numerical stability.
We then compute a final-answer advantage and a step-specific memory advantage:
\begin{align}
    A_g^\text{ans} = \text{GN}(r_g^a), \quad
    A_{g,t}^\text{mem} = \begin{cases}
    A_g^\text{ans} + \alpha\text{GN}(r_{g,t}^{m}), & t<T, \\
    A_g^\text{ans}, & t=T,
    \end{cases}
\end{align}
where normalization of $r_{g,t}^m$ is performed across the $G$ rollouts at the same step $t$, and $\alpha$ controls the strength of next-preference supervision.
Thus, every memory is credited for the final outcome, while non-terminal memories additionally receive immediate feedback about whether they predict the next preference correctly.

\subsubsection{\textbf{Memory-Level Policy Ratio}}
For the $t$-th intermediate memory in the $g$-th sampled instance, we define the token-level policy ratio at $l$-th token as
\begin{equation}
    \rho^{m}_{g,t,l}(\theta)
    =
    \frac{
    \pi_{\theta}
    \left(
    m_{g,t,l}
    \mid \mathcal{Q}, \mathbf{m}_{g,t,<l}, \mathbf{m}_g^{t-1},\mathbf{c}^t
    \right)
    }{
    \pi_{\theta_{\mathrm{old}}}
    \left(
    m_{g,t,l}
    \mid \mathcal{Q}, \mathbf{c}_{g,t,<l}, \mathbf{m}_g^{t-1},\mathbf{c}^t
    \right)
    }.
\end{equation}
Each memory is conditioned only on the original query $x$, the given chunk $\mathbf{c}^t$ and its own previous tokens $\mathbf{m}_{g,t,<l}$, rather than on tokens from other memory summaries.
This prevents artificial cross-memory credit assignment.
Similarly, the token-level policy ratio of final output can be formulated as
\begin{equation}
    \rho^{y}_{g,l}(\theta)
    =
    \frac{
    \pi_{\theta}
    \left(
    y_{g,l}
    \mid \mathcal{Q}, \mathbf{m}^T_g, y_{g,<l}, \hat{\mathbf{c}}
    \right)
    }{
    \pi_{\theta_{\mathrm{old}}}
    \left(
    y_{g,l}
    \mid \mathcal{Q}, \mathbf{m}^T_g, y_{g,<l}, \hat{\mathbf{c}}
    \right)
    }.
\end{equation}

\subsubsection{\textbf{Multi-Memory GRPO Objective}}
Let
\begin{align}
        \ell_{\text{clip}}(\rho,A)
    =\min (
    \rho A,
    \text{clip}(\rho,1-\delta,1+\delta) A),
\end{align}
where $\delta$ is the clipping coefficient (0.2). The answer and memory objectives are
\begin{align}
    \mathcal{J}_{\mathrm{Ans}}(\theta)
    &=\mathbb{E}\left[
    \frac{1}{G}\sum_{g=1}^G
    \frac{1}{L_y}\sum_{l=1}^{L_y}
    \ell_{\mathrm{clip}}
    \big(\rho^y_{g,l}(\theta),A_g^{\mathrm{ans}}\big)
    \right], \nonumber\\
    \mathcal{J}_{\mathrm{Mem}}(\theta)
    &=\mathbb{E}\left[
    \frac{1}{G}\sum_{g=1}^G
    \frac{1}{T}\sum_{t=1}^T
    \frac{1}{L_m}\sum_{l=1}^{L_m}
    \ell_{\mathrm{clip}}
    \big(\rho^m_{g,t,l}(\theta),A_{g,t}^{\mathrm{mem}}\big)
    \right].
    \label{eq:mm_grpo_components}
\end{align}

The resulting objective is
\begin{equation}   \mathcal{J}_{\mathrm{Mem\text{-}GRPO}}(\theta)
    =
    \mathcal{J}_{\mathrm{Ans}}(\theta) 
    +
    \lambda \mathcal{J}_{\mathrm{Mem}}(\theta),
\end{equation}
where $\lambda$ controls whether and how strongly the memory tokens are optimized.

\subsubsection{\textbf{KL regularization}}
Because the summary component $\mathbf{s}_g^t$ lacks direct intermediate supervision, we introduce a token-level KL penalty over the generated memory tokens to prevent the learned policy $\pi_\theta$ from deviating excessively from a reference policy $\pi_{\mathrm{ref}}$:
\begin{align}
\begin{aligned}
    \mathcal{D}_{\mathrm{KL}}^{m}
    & =
    \mathbb{E}
    \Bigg[
    \frac{1}{G}
    \sum_{g=1}^{G}
    \frac{1}{T}
    \sum_{t=1}^{T}
    \frac{1}{L_m}
    \sum_{l=1}^{L_m} (\pi - \log \pi -1)
    \Bigg], \\
    & \text{where} \quad  \pi = 
    \frac{
    \pi_{\mathrm{ref}}
    \left(
    m_{g,t,l}
    \mid \mathcal{Q}, m_{g,t,<l}, \mathbf{c}^t
    \right)
    }{
    \pi_{\theta}
    \left(
    m_{g,t,l}
    \mid \mathcal{Q}, m_{g,t,<l}, \mathbf{c}^t
    \right)
    }.
\end{aligned}
\end{align}

The final training objective is
\begin{equation}
    \max_\theta \mathcal{J}(\theta)
    =
    \mathcal{J}_{\mathrm{Mem\text{-}GRPO}}(\theta)
    -
    \beta \mathcal{D}_{\mathrm{KL}}^{m},
\end{equation}
where $\beta$ is the KL coefficient (set to 0.04 following~\cite{shao2024deepseekmath}).
We do not introduce an additional reference-policy KL penalty for answer tokens because they already receive direct supervision through the formatting and final-answer rewards.

\section{Experiment}
\subsection{Experimental Settings}
\subsubsection{\textbf{Datasets}}

To evaluate the effectiveness of the proposed \ourname{} in the user-agent-platform paradigm with proactive user instructions, we construct three agentic recommendation datasets, namely MovieTV, Books, and Games using the widely-used \textbf{\emph{Amazon Reviews 2023}}\footnote{\url{https://amazon-reviews-2023.github.io/}} data, which provide textual information (such as item titles, descriptions, and reviews), as well as the URL links of product images.
Based on these contents, we create item snapshots into a format of shopping platform webpage, and randomly include advertising content as a distraction.
Furthermore, following InstructRec~\cite{zhang2026recommendation}, we assign a persona to each user and generate the instruction for this interaction based on the corresponding user review and task.
More details are in the Appendix~\ref{app:data}.

\begin{table*}[htbp]
  \centering
  \captionsetup{font={small}}
  \caption{Basic statistics of benchmark datasets.}
  \scalebox{0.85}{
    \begin{tabular}{c|ccccc}
    \toprule
    \textbf{Dataset} & \textbf{\#Users} & \textbf{\#Items} & \textbf{\#Int.} & \textbf{Avg. Seq.} & \textbf{\#Tokens (Avg. | Max.)} \\
    \midrule
    Games & 94,762 & 25,612 & 570,720 & 8.7764 & 26,217.33 | 722,968 \\
    Books & 7,377  & 120,925 & 207,759 & 28.1631 & 62,281.48 | 5,972,639 \\
    MovieTV & 5,649  & 28,987 & 79,737 & 14.1166 & 30,207.68 | 292,389 \\
    \bottomrule
    \end{tabular}%
    }
  \label{tab:dataset}%
\end{table*}%

\subsubsection{\textbf{Compared Models}}
We include four classes of baselines: (i) conventional sequential recommendation methods, namely SASRec~\cite{kang2018self} and BERT4Rec~\cite{sun2019bert4rec};
(ii) natural language assists for recommendation: P5~\cite{geng2022recommendation} and TokenRec~\cite{qu2025tokenrec};
(iii) recommendation agents: ToolRec~\cite{zhao2024let} and iAgent~\cite{xu2025iagent};
(iv) long-context LLM agents: QwenLong-L1 (Qwen-L1)~\cite{wan2025qwenlong}, Mem0~\cite{chhikara2025mem0}, and MemAgent~\cite{yu2026memagent}.

\subsubsection{\textbf{Configurations}}
We implement all models using Python 3.12 and Unsloth\footnote{\url{https://unsloth.ai/}} (Version 2026.5.2) on four NVIDIA H20 (96 GB) GPU.
For controlled evaluation, we model the workflow as chronological item-click actions.
Calls are separately instantiated but recurrently linked.
Evaluation is offline and history-conditioned; rejection, clarification, exposure, and behavioral feedback remain outside scope.
We implement three evaluation tasks for recommendation agents: (i) \emph{Searching}, which retrieves the proper item from the candidate set; (ii) \emph{Ranking}, which sorts the candidate set; and (iii) \emph{Judging}, which determines whether the user would like a given item.
For each sample, we randomly select 9 negative items and combine them with the target item to form a candidate set~\cite{xu2025iagent,shi2025personax,zhang2026recommendation}.
Taking into account both performance and computational resources, we opt to utilize the \emph{Qwen3.5-9B} as our LLM backbone, which is renowned as one of the most popular pretrained vision-language models globally.
The LLM backbone operates with a low temperature of 0.2 for inferencea and a high temperature of 1.5 for the GRPO training.
The maximum length of the item sequence is configured to 50 to embody long-horizon user preference evolving.
Our memory length $L_m$ is configured to $5,000$ language tokens, while the chunking window size $W$ is set to 3, and the maximum length of each chunk $C$ can be upto $100,000$ tokens.
Consequently, the model typically requires 3 to 5 conversational turns to process the entire context.
The weight of next-preference supervision $\alpha$ is set to 0.5.
The loss weight $\lambda$ is set to 0 in the two epochs (warming up), and set to $0.7$ for the following training.
We use a rollout batch size of 64 and a group size of 8 for our training.

\subsubsection{\textbf{Evaluation Metrics}}
Three commonly used metrics are used: item-title matching Accuracy (Acc.) for the \emph{Judging} task, Top-K Hit Ratio ({H@K}) and Top-K Normalized Discounted Cumulative Gain ({N@K}) for the \emph{Searching} and \emph{Ranking} tasks, respectively, where higher values indicate superior performance.
Furthermore, the values of K are specified as 1, 3 and 5, with 1 and 3 serving as the default settings of HR and NDCG, respectively, for ablation experiments and parameter analyses.

\subsection{Recommendation Performance}

\begin{table*}[htbp]
  \centering
  \captionsetup{font={small}}
  \caption{Performance comparison between representative methods across three commonly used datasets on three recommendation tasks. The best and second-best results are highlighted in bold and underlined fonts, respectively. For \ourname{}, we conduct independent inference five times and report the mean and standard deviation in separate rows. The improvements over baselines are statistically significant ($p<0.01$).}
  \begin{threeparttable}
  \resizebox{\textwidth}{!}{
    \begin{tabular}{cc|cccccc|ccc|ccc}
    \toprule
    \multirow{3}{*}{\textbf{Category}}
    & \multirow{3}{*}{\textbf{Method}}
    & \multicolumn{6}{c|}{\textbf{Searching}}
    & \multicolumn{3}{c|}{\textbf{Ranking}}
    & \multicolumn{3}{c}{\textbf{Judging}} \\
    
    \cmidrule(lr){3-8}
    \cmidrule(lr){9-11}
    \cmidrule(lr){12-14}
    
    &
    & \multicolumn{2}{c}{\textbf{Games}}
    & \multicolumn{2}{c}{\textbf{MovieTV}}
    & \multicolumn{2}{c|}{\textbf{Books}}
    & \textbf{Games}
    & \textbf{MovieTV}
    & \textbf{Books}
    & \textbf{Games}
    & \textbf{MovieTV}
    & \textbf{Books} \\
    
    &
    & \textbf{H@1}
    & \textbf{H@3}
    & \textbf{H@1}
    & \textbf{H@3}
    & \textbf{H@1}
    & \textbf{H@3}
    & \textbf{N@3}
    & \textbf{N@3}
    & \textbf{N@3}
    & \textbf{Acc.}
    & \textbf{Acc.}
    & \textbf{Acc.} \\
    
    \midrule
    
    \multirow{2}{*}{\textbf{DeepRec}}
    & \textbf{SASRec}
    & 0.2828
    & 0.3115
    & 0.1763
    & 0.2208
    & 0.1761
    & 0.2853
    & 0.2689
    & 0.2605
    & 0.2261
    & 0.6806
    & 0.6004
    & 0.6431 \\
    
    & \textbf{BERT4Rec}
    & 0.2955
    & 0.3225
    & 0.1838
    & 0.2320
    & 0.1898
    & 0.2701
    & 0.2727
    & 0.2268
    & 0.2288
    & 0.6739
    & 0.5933
    & 0.6391 \\

    \multirow{2}{*}{\textbf{LLMRec}}
    & \textbf{P5}
    & 0.2509
    & 0.2525
    & 0.1319
    & 0.1816
    & 0.1411
    & 0.2248
    & 0.2815
    & 0.2470
    & 0.2464
    & 0.6775
    & 0.6120
    & 0.6560 \\
    
    & \textbf{TokenRec}
    & 0.3011
    & 0.3711
    & 0.1799
    & 0.2501
    & 0.2232
    & 0.3228
    & 0.3891
    & 0.2985
    & 0.3107
    & 0.6997
    & 0.6456
    & 0.7142 \\

    \multirow{2}{*}{\textbf{RecAgent}}
    & \textbf{ToolRec}
    & 0.1970
    & 0.3636
    & 0.1613
    & 0.1827
    & 0.2172
    & 0.3831
    & 0.3969
    & 0.3165
    & 0.2594
    & 0.7296
    & 0.6737
    & 0.7246 \\
    
    & \textbf{iAgent}
    & 0.3824
    & 0.5096
    & 0.2308
    & \underline{0.3536}
    & 0.2925
    & \underline{0.4857}
    & \underline{0.4816}
    & \underline{0.3817}
    & \underline{0.3839}
    & 0.7514
    & 0.6799
    & 0.7138 \\

    \multirow{3}{*}{\textbf{LongAgent}}
    & \textbf{Qwen-L1}
    & 0.3249
    & 0.4851
    & 0.2260
    & 0.3162
    & 0.2843
    & 0.3999
    & 0.4368
    & 0.3334
    & 0.3360
    & 0.7781
    & 0.6677
    & \underline{0.7466} \\
    
    & \textbf{Mem0}
    & \underline{0.3995}
    & \underline{0.5184}
    & 0.2480
    & 0.3342
    & 0.3133
    & 0.4756
    & 0.4616
    & 0.3269
    & 0.3618
    & \underline{0.7799}
    & 0.6933
    & 0.7206 \\

    & \textbf{MemAgent}
    & 0.3864
    & 0.5017
    & \underline{0.2499}
    & 0.3370
    & \underline{0.3172}
    & 0.4628
    & 0.4620
    & 0.3296
    & 0.3629
    & 0.7714
    & \underline{0.6960}
    & 0.7304 \\

    \midrule
    
    \multirow{2}{*}{\textbf{\ourname{}}}
     & \textbf{Mean}
     & \textbf{0.4449}  
     & \textbf{0.5753} 
     & \textbf{0.2610} 
     & \textbf{0.3740} 
     & \textbf{ 0.3663} 
     & \textbf{0.5133} 
     & \textbf{0.5222} 
     & \textbf{0.3863} 
     & \textbf{0.4226} 
     & \textbf{0.8350} 
     & \textbf{0.7364} 
     & \textbf{0.7818} \\
    
    & \textbf{Std}
    & 0.0252
    & 0.0384
    & 0.0168
    & 0.0415
    & 0.0198
    & 0.0501
    & 0.0341
    & 0.0268
    & 0.0199
    & 0.0080
    & 0.0079
    & 0.0086 \\
    
    \midrule
    
    \multicolumn{2}{c|}{\textbf{Improvement*}}
    & 11.37\% & 10.97\% & 4.42\% & 5.77\% & 15.49\% & 5.69\% & 8.43\% & 1.21\% & 10.08\% & 7.06\% & 5.80\% & 4.72\%
    \\
    
    \bottomrule
    \end{tabular}
  }
  \begin{tablenotes}
    \footnotesize
    \item[1] * denotes the improvement over the strongest baseline.
    \item[2] Mean and Std denote the mean and standard deviation over five
    independent inference runs, respectively.
  \end{tablenotes}
  \end{threeparttable}
  \label{tab:comparison}
\end{table*}

Table~\ref{tab:comparison} presents the overall performance comparison, from which we make the following observations:

\begin{itemize}[leftmargin=*]
\item {\ourname{} achieves the best overall performance across most datasets and tasks.}
\ourname{} obtains the best results on all searching and judging settings, and achieves the best performance on two out of three ranking settings.
On average, \ourname{} consistently improves over the strongest baseline by \improve\%.
These consistent gains demonstrate the effectiveness of \ourname{} in identifying user-preferred items from candidate sets.
The improvements are statistically significant with $p<0.01$, indicating that the advantage of our model is stable rather than incidental.
Moreover, the standard deviations of \ourname{} are relatively small, demonstrating stable inference behavior.

\item {Existing RecAgent baselines improve over conventional recommendation methods but remain limited.}
Compared with DeepRec methods such as SASRec and BERT4Rec, and LLMRec methods such as P5 and TokenRec, agent-based methods generally obtain stronger performance, especially in searching and judging tasks.
For instance, ToolRec and iAgent achieve competitive results on several datasets, suggesting that interactive reasoning and tool-augmented decision-making are beneficial for recommendation agents.
However, these methods are still consistently outperformed by \ourname{}.
This indicates that equipping an LLM with tools or interactive conversations is insufficient for complex agentic recommendation scenarios.
Without reliable item perception and efficient memory updating, agents may still suffer from noisy observations, redundant histories, and degraded reasoning over long interaction traces.

\item {Long-context modeling alone does not guarantee better recommendation performance.}
Although LongAgent baselines, such as Qwen-L1, Mem0, and MemAgent, are designed to handle extended contexts, their performance is not consistently superior to RecAgent methods, in particularly for the \emph{Ranking} task.
For example, in the ranking task on MovieTV, Mem0 obtains 0.3269 NDCG@3, which is lower than iAgent's 0.3817 and also lower than \ourname{}.
This suggests that directly extending the context window or maintaining coarse memory may introduce irrelevant historical information and increase reasoning difficulty.
In contrast, \ourname{} tends to preserve essential preference information and maintains effective memory, leading to more accurate recommendation.

\end{itemize}

\subsection{In-depth Analysis}

\subsubsection{\textbf{Various-Length Reasoning}}

\begin{figure*}[h]
    \centering
    \includegraphics[width=\linewidth]{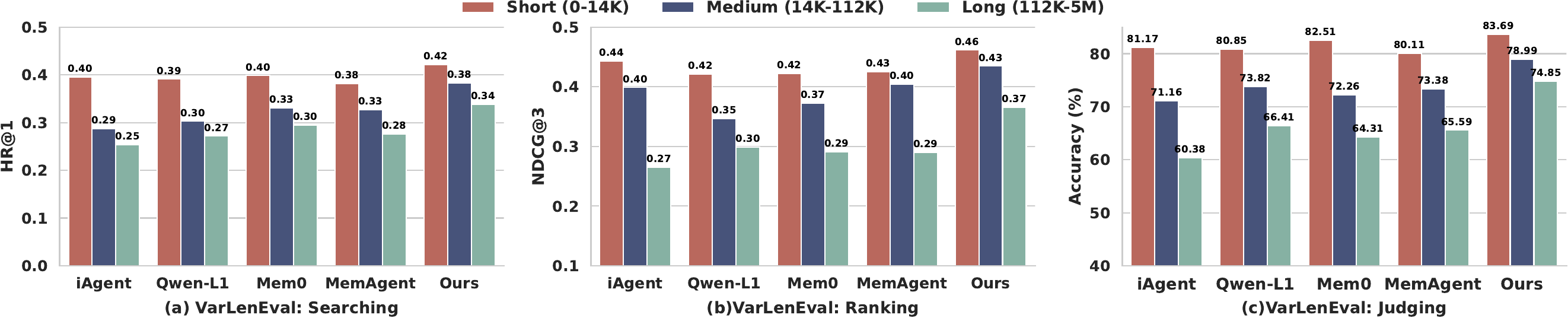}
      \captionsetup{font={small}}
    \caption{Analysis of reasoning capability across contexts of varying lengths on the \emph{Books} dataset for three recommendation tasks.}
    \label{fig:varlen}
\end{figure*}

This section examines the ability of our proposed model to capture user preferences from contexts of varying lengths.
We divide the evaluation samples into three context horizons: short (0--14K tokens), medium (14K--112K tokens), and long (over 112K tokens).
We report the results on the \emph{Books} dataset, which contains 658 short-context samples, 6,024 medium-context samples, and 694 long-context samples.
As shown in Figure~\ref{fig:varlen}, all models exhibit a performance decline as the context length increases, indicating that long-horizon preference modeling remains challenging for recommender agents. Nevertheless, our model consistently achieves the best performance across the three tasks and all context-length groups.
Notably, our model’s performance under long-context settings remains comparable to, or even better than, several baselines under medium-context settings.
This suggests that the proposed dynamic memory mechanism effectively filters, organizes, and retrieves preference-relevant information from lengthy contexts, reducing the negative impact of irrelevant or noisy historical interactions.
These results verify that our model is better suited for long-horizon recommendation reasoning, where accurately capturing evolving and sparse user preferences is essential.

\subsubsection{\textbf{Ablation Study}}

We report the results of ablation experiments on the \emph{Searching} task, where three key modules are removed separately as follows:
\begin{itemize}[leftmargin=*]
\item - \textit{MM GRPO}: The Multi-Memory GRPO module is replaced by standard GRPO using only the final-answer reward.
\item - \textit{OCR}: The vision-based OCR perception module is deactivated, and raw item contents are used instead.
\item - \textit{RL}: The GRPO post-training is removed, resulting in a training-free variant of the proposed model.
\item - \textit{TEM}: The Time-Evolving Memory (TEM) is replaced with a plain long-text context consisting of interaction history, item information, and user instructions.
\end{itemize}

Table~\ref{tab:ablation} shows that removing any module consistently degrades performance across all datasets, confirming that each component contributes to the effectiveness of \ourname{}.
Replacing MM GRPO with standard GRPO consistently reduces accuracy, demonstrating the benefit of memory-specific rewards.
Disabling {OCR} causes clear drops, particularly on Games and Books, confirming the value of structured visual perception.
Removing {RL} further degrades performance, although the training-free variant remains competitive with strong baselines such as iAgent.
Removing {TEM} causes the largest decline across all datasets and metrics, suggesting that a plain long context introduces irrelevant information and complicates preference reasoning.
In contrast, TEM enables more effective modeling of evolving preferences.
Overall, OCR, Multi-Memory GRPO, and TEM play complementary roles in item perception, memory optimization, and long-context preference modeling.

\begin{table*}[h]
\centering
  \captionsetup{font={small}}
    \caption{Results of Ablation Studies on the \emph{Searching} task.}
    \scalebox{0.85}{
    \begin{tabular}{ccccccc}
        \toprule
       \multirow{2}{*}{Module} & \multicolumn{2}{c}{{Games}} & \multicolumn{2}{c}{{MovieTV}} & \multicolumn{2}{c}{{Books}} \\ \cmidrule(lr){2-3} \cmidrule(lr){4-5} \cmidrule(lr){6-7}
        & {HR@1}      & {HR@3}      & {HR@1}      & {HR@3}      & {HR@1}      & {HR@3}      \\ \midrule
        \textbf{\ourname{} (Full)} &  \textbf{0.4449} & \textbf{0.5753} & \textbf{0.2610} & \textbf{0.3740} & \textbf{0.3663} & \textbf{0.5133}  \\
        \quad - \textit{MM GRPO} & 0.4292 & 0.5590 & 0.2552 & 0.3656 & 0.3588 & 0.5069 \\
        
        \quad - \textit{OCR} & 0.4072 & 0.5144 & 0.2543 & 0.3593 & 0.3231 & 0.4980 \\
        \quad - \textit{RL} & 0.3816 & 0.5200 & 0.2482 & 0.3535 & 0.2841 & 0.4864 \\
       \quad - \textit{TEM} & 0.3465 & 0.5009 & 0.2347 & 0.3269 & 0.2593 & 0.4139   \\
        \bottomrule
    \end{tabular}
    }
    \label{tab:ablation}
\end{table*}

\section{Conclusion}
\label{sec:conclusion}
In this paper, we propose \ourname{}, a RecAgent framework that improves both item perception and long-context preference modeling.
\ourname{} replaces brittle HTML parsing with OCR-based perception, enabling multimodal extraction of structured item information from screenshots.
It further introduces a chunk-wise dynamic memory mechanism that maintains informative user histories with bounded context length and linear inference complexity.
To better train memory updates, we design a GRPO variant that leverages recommendation-specific information and evaluation metrics.
This work is the first to explore perception and long-context reasoning for RecAgents. Experiments across diverse recommendation tasks show that \ourname{} achieves superior accuracy and efficiency compared with representative baselines.

\bibliographystyle{IEEEtran}
\bibliography{references}

\clearpage
\appendix

\noindent \textbf{Appendix Outline}
\begin{itemize}

    \appoutlineitem{app:literature}
    {Related Work}

    \begin{itemize}[leftmargin=1.5em, label=--, itemsep=0.05em, topsep=0.1em]
        \appoutlineitem{app:llmagent}
        {LLM Agents}

        \appoutlineitem{app:ocr}
        {OCR-based Perception}

        \appoutlineitem{app:long}
        {Long-Context Reasoning}

        \appoutlineitem{app:disctinction}
        {Distinction from Existing Studies}
    \end{itemize}

    \appoutlineitem{app:method}
    {Supplements on Research Methods}

    \begin{itemize}[leftmargin=1.5em, label=--, itemsep=0.05em, topsep=0.1em]
        \appoutlineitem{app:GRPO}
        {Standard GRPO}

        \appoutlineitem{app:reward}
        {Task-specific Reward Modeling}

        \appoutlineitem{app:template}
        {Prompt Templates}
    \end{itemize}

    \appoutlineitem{app:setting}
    {Additional Details about Experiments}

    \begin{itemize}[leftmargin=1.5em, label=--, itemsep=0.05em, topsep=0.1em]
        \appoutlineitem{app:data}
        {Supplements on Datasets}

        \appoutlineitem{app:baseline}
        {Baselines}

        \appoutlineitem{app:ocr_eval}
        {OCR Evaluation}

        \appoutlineitem{app:case}
        {Case Study}

        \appoutlineitem{app:VarLenEval}
        {Supplements on Various-Length Reasoning}

        \appoutlineitem{app:param}
        {Hyperparameter Analysis}

        \appoutlineitem{app:latency}
        {Inference Latency Analysis}
    \end{itemize}

    \appoutlineitem{app:future_work}
    {Discussion on Limitations and Future Work}

\end{itemize}

\section{Related Work}
\label{app:literature}

\subsection{LLM Agents}
\label{app:llmagent}
LLM agents combine planning, memory, reflection, and tool use for tasks such as deep research~\cite{zhao2024expel}, scientific discovery~\cite{schmidgall2025agent,dai2025large}, and web interaction~\cite{ning2025survey}.
Recommendation agents (RecAgents) further use LLMs to interpret intent, model preferences, invoke tools, and interactively recommend items~\cite{shi2024large,zhang2024generative,zhang2024agentcf,xu2025iagent,liu2026recoworld,shi2025personax,huang2025recommender,wang2024recmind,huang2026towards}.
However, most RecAgents assume structured or textual item evidence and implement memory through prompts, profiles, or external components, leaving page-level perception and histories beyond a single context window insufficiently explored.

\subsection{OCR-based Perception}
\label{app:ocr}
Modern OCR systems, including PaddleOCR~\cite{cui2025paddleocr}, DeepSeek-OCR~\cite{wei2025deepseek}, and OCR-2~\cite{wei2026deepseek}, recover textual and structural evidence from screenshots and user interfaces.
For RecAgents, OCR can normalize heterogeneous rendered pages into LLM-readable item evidence without requiring platform-specific schemas.
It can expose item cards, prices, descriptions, comments, and other interface-level signals from pages designed primarily for human perception.
Although web agents use screenshots or textualized interfaces~\cite{gur2024real,he2024webvoyager}, OCR-based page perception remains underexplored in recommendation; here, OCR serves as an evidence extractor rather than a substitute for general visual reasoning.

\subsection{Long-Context Reasoning}
\label{app:long}
Long and noisy contexts remain challenging for LLM agents~\cite{chhikara2025mem0,hsieh2024ruler,liu2024lost,wang2026vtc}, particularly when item pages and interaction histories exceed the context window~\cite{qu2025tokenrec,qu2026diffusion}.
Prior solutions either process an extended context directly~\cite{wan2025qwenlong}, retrieve from external stores or profiles~\cite{chhikara2025mem0,liu2025agentcf++,xu2026mem}, or recurrently update a compact state over textual chunks~\cite{yu2026memagent,zhou2026mem}.
The first incurs growing context cost, the second requires auxiliary infrastructure, and the third must learn what to retain.
Ideally, a long-context agent should process unbounded input, avoid performance degradation as the context grows, and retain efficient decoding with linear processing complexity~\cite{yu2026memagent}.
This goal mirrors human note-taking: abstract central concepts and retain critical details while discarding redundant information.
We adopt the recurrent design but specialize its state and learning objective for evolving recommendation preferences.

\subsection{Distinction from Existing Studies}
\label{app:disctinction}
Existing RecAgents and memory agents address complementary parts of our problem; \ourname{} connects page perception, recurrent preference memory, and recommendation specific memory learning in one framework (Table~\ref{tab:literature_comparison}).
iAgent~\cite{xu2025iagent} models user interests but assumes structured or textual observations; Mem0~\cite{chhikara2025mem0} relies on external storage and retrieval; and MemAgent~\cite{yu2026memagent} and MEM1~\cite{zhou2026mem} learn recurrent memory states for generic long-document reasoning.
In contrast, \ourname{} (i) parses rendered item pages with OCR, (ii) maintains a bounded, human-readable state representing evolving preferences, and (iii) trains memory updates with both final-recommendation and intermediate next-preference rewards through Multi-Memory GRPO.
With fixed chunk and memory budgets, this design keeps the active context bounded and scales linearly with the number of chunks without an external retrieval system or backbone modification.
To the best of our knowledge, \ourname{} is the first RecAgent to combine these three capabilities across searching, ranking, and judging tasks; our novelty claim concerns their recommendation-specific integration rather than any individual component.
The design follows a deliberately simple principle: long-context preference modeling remains inside the standard autoregressive process, without an external retrieval system, a specialized memory module, or a backbone architecture change.

\begin{table*}[h]
    \centering
    \captionsetup{font={small}}
    \caption{Comparison with closely related agent designs. Symbols indicate whether each capability is explicitly supported.}
    \label{tab:literature_comparison}
    \setlength{\tabcolsep}{1.5pt}
    \renewcommand{\arraystretch}{1.12}
    \scalebox{0.85}{
    \begin{tabularx}{\columnwidth}{@{}ll>{\raggedright\arraybackslash}Xccc@{}}
        \toprule
        \textbf{Method} & \textbf{Target} & \textbf{Memory Form}
        & \shortstack{\textbf{Multimodal}\\\textbf{perception}}
        & \shortstack{\textbf{Readable}\\\textbf{memory}}
        & \shortstack{\textbf{Rec.-specific}\\\textbf{memory RL}} \\
        \midrule
        iAgent~\cite{xu2025iagent}
        & Rec
        & Simulate user behavior
        & \featno & \featno & \featno \\

        Mem0~\cite{chhikara2025mem0}
        & General
        & External store and retrieval
        & \featno & \featno & \featno \\

        MemAgent~\cite{yu2026memagent}
        & Long
        & Recurrent token memory
        & \featno & \featyes & \featno \\
        
        \midrule
        \textbf{\ourname{}}
        & {Rec}
        & {Recurrent preference memory}
        & \featyes & \featyes & \featyes \\
        \bottomrule
    \end{tabularx}
    }
\end{table*}

\section{Supplements on Research Methods}
\label{app:method}

\subsection{Standard GRPO}
\label{app:GRPO}

To better contextualize the proposed Multi-Mem GRPO, this section introduces the standard formulation of Group Relative Policy Optimization (GRPO) as a baseline for comparison.
Given a prompt $x$, GRPO samples a group of $G$ responses
$\{y_g\}_{g=1}^{G}$ from the old policy $\pi_{\theta_{\mathrm{old}}}$:
\begin{align}
    y_g \sim \pi_{\theta_{\mathrm{old}}}(\cdot \mid x), \quad g=1,\dots,G.
\end{align}
Each response $y_g$ receives a scalar reward $r_g = r(x,y_g)$. Instead of using a learned value function, GRPO estimates the advantage by normalizing rewards within the sampled group:
\begin{align}
    \hat{A}_g
    =
    \frac{r_g - \mathrm{mean}(\{r_g\}_{g=1}^{G})}
    {\mathrm{std}(\{r_g\}_{g=1}^{G}) + \epsilon},
\end{align}
where $\epsilon$ is a small constant for numerical stability.

The policy is optimized with a PPO-style clipped objective and a KL regularization term against a reference policy $\pi_{\mathrm{ref}}$:
\begin{align}
\begin{aligned}
\mathcal{J}_{\mathrm{GRPO}}(\theta)
& =
\mathbb{E}_{x, \{y_g\}_{g=1}^{G}}
[
\frac{1}{G}
\sum_{g=1}^{G}
\frac{1}{|y_g|}
\sum_{l=1}^{|y_g|} \\
& \min
(
\rho_{g,l}(\theta)\hat{A}_g,
\mathrm{clip}\big(\rho_{i,l}(\theta), 1-\varepsilon, 1+\varepsilon\big)\hat{A}_g
) \\
& -
\beta \, D_{\mathrm{KL}}
(
\pi_{\theta}(\cdot \mid x)
\,\|\, 
\pi_{\mathrm{ref}}(\cdot \mid x)
)
],
\end{aligned}
\end{align}
where
\begin{align}
    \rho_{g,l}(\theta)
    =
    \frac{
    \pi_{\theta}(y_{g,l} \mid x, y_{g,<l})
    }{
    \pi_{\theta_{\mathrm{old}}}(y_{g,l} \mid x, y_{g,<l})
    },
\end{align}
$\varepsilon$ is the clipping threshold, $\beta$ controls the strength of KL regularization, and $D_{\mathrm{KL}}(\cdot\|\cdot)$ penalizes deviation from the reference policy.

Equivalently, the training objective is to maximize:
\begin{align}
    \theta^{*}
    =
    \arg\max_{\theta}
    \mathcal{J}_{\mathrm{GRPO}}(\theta).
\end{align}

\subsection{Reward Modeling}
\label{app:reward}
The format reward $r_f$ provides graded supervision for adherence to the required XML-style response structure, following the rule-based format checks used in the Unsloth GRPO implementation.
In our framework, this reward is instantiated as $r_f^m$ for intermediate memory outputs and $r_f^a$ for final-answer outputs.
It consists of three additive terms:
\begin{align}
     r_f = r_{\mathrm{count}} + r_{\mathrm{soft}} + r_{\mathrm{strict}}.   
\end{align}
First, the XML-count term $r_{\mathrm{count}}$ assigns 0.125 points for each correctly placed structural marker.
Intermediate memory outputs use \texttt{<memorize>}, \texttt{</memorize>}, \texttt{<predict>}, and \texttt{</predict>}, whereas final-answer outputs use \texttt{<answer>} and \texttt{</answer>}.
The maximum XML-count reward is therefore 0.5.
Second, the soft-format term $r_{\mathrm{soft}}$ assigns 0.5 points when the two required blocks appear in the prescribed order: a \texttt{memorize} block followed by a \texttt{predict} block for intermediate memory, or an \texttt{answer} block for the final output.
Third, the strict-format term $r_{\mathrm{strict}}$ assigns an additional 0.5 points only when the complete response exactly matches the corresponding multiline template.
Thus, $r_f \in [0,1.5]$, with a maximum of 1.5 awarded to a fully compliant response.

\subsection{Prompt Templates}
\label{app:template}
In this section, we list out all the prompt templates in our framework. 
In which, curly-brace placeholders ${}$ will be replaced with actual content.
First, there are two prompts, "TEMPLATE\_MEMORY" and "TEMPLATE\_FINAL", which designed to memory processing (top) and final answer generation (bottom).

\begin{tcolorbox} 
TEMPLATE\_MEMORY = """
\#\#\# You are presented with a problem, a section of an article that may contain the answer to the problem, and a previous memory. Please read the provided section carefully and update the memory with the new information that helps to answer the problem. Be sure to retain all relevant details from the previous memory while adding any new, useful information. Building upon the memory, please make a simple prediction of user's preference of item characteristics.

<problem> 
\colorbox{cyan!30}{\{query\}}
</problem>

<memory>
\colorbox{yellow!30}{\{memory\}}
</memory>

<section>
\colorbox{orange!30}{\{chunk\}}
</section>

\

\#\#\# Updated memory:

<memorize>
\colorbox{gray!30}{\{intermediate memory\}}
</memorize>

<predict>
\colorbox{gray!30}{\{intermediate prediction\}}
</predict>
"""

\tcblower
TEMPLATE\_FINAL = """
\#\#\# You are presented with a problem and a previous memory. Please answer the problem based on the previous memory and put the answer in boxed\{\{$\cdot$\}\}.

<problem> 
\colorbox{cyan!30}{\{query\}} + \colorbox{orange!30}{\{candidates\}}
</problem>

<memory>
\colorbox{yellow!30}{\{memory\}}
</memory>

\#\#\# Your Answer:

<answer>
\colorbox{gray!30}{\{\{target item title\}\}}
</answer>
"""
\end{tcolorbox}

\begin{tcolorbox} 
QUERY\_SEARCHING = "Now, what would be the next possible item for the user from the candidates based on the user's interaction history and instruction? Please just give the title of the item as the answer; no explanation is needed."
\end{tcolorbox}

\begin{tcolorbox} 
QUERY\_RANKING = "You are given a set of candidate items. Rank the products according to how likely the user is to prefer them based on the user's interaction history and instruction. Return only the product titles in ranked order, with no additional explanation."
\end{tcolorbox}

\begin{tcolorbox} 
QUERY\_JUDGING = "Now, is the user likely to interact with the given item? Please answer with a single word: 'Yes' or 'No'. No explanation is needed."
\end{tcolorbox}

\begin{tcolorbox}
LLM-AS-JUDGE = """You are a general AI assistant.

\

Based on the [Correct Answer] provided below, determine whether the [Response] to the [Original Question] is correct.

[Original Question]: \{question\}

[Correct Answer]: 
\{golden\_answer
\}

[Response]: 
\{pred\_answer
\}

\

Your judgment must follow this standard:

- Focus only on whether there are substantial differences between the [Response] and the [Correct Answer]

- Do not comment on the background of the question

- Do not attempt to resolve the problem again

- Only focus on judging whether the answers are consistent

- If the [Response] is consistent with the [Correct Answer], or within an acceptable small margin of error for numerical questions, judge as "correct"

- Otherwise (i.e., in cases of any inconsistency, ambiguity, non-equivalence, or incorrectly extracted answer), judge as "incorrect"

\

Output JSON format:

\{\{
  "judgement": "correct" or "incorrect"
\}\}"""

\end{tcolorbox}

\section{Additional Details about Experiment}
\label{app:setting}

\subsection{Supplements on Datasets}
\label{app:data}

\noindent\textbf{WebWalkerQA}~\cite{wu2025webwalker}. The dataset is in the form of a JSON file with a collection of 680 questions and answers.
However, as time passing by, part of the websites had been updated.
Thus, we removed the invalid questions and updated the outdated answers after a human-verification, with a collection of 673 samples remains.
The task is to access the website and retrieve useful information to answer the question.
The average length of the textual contents on the websites is around 716.41 tokens.
A representative example of a WebWalkerQA sample is list as below:
\begin{itemize} [leftmargin=*]
    \item "\emph{question}": "How many CCF-B level conference papers did Associate Professor XXX from the XXX University School of Computer Science publish between 2021 and 2023?"
    \item "\emph{answer}": "4"
    \item "\emph{root\_url}": [URL]
    \item "\emph{source\_website}": [URL]
    \item "\emph{golden\_path}": [classified]
    \item "\emph{difficulty\_level}": "medium"
\end{itemize}

\noindent\textbf{HotpotQA}~\cite{yang2018hotpotqa}.
HotpotQA is a new dataset with 113k Wikipedia-based question-answer pairs.
In this study, we use a long-context synthetic variant of the HotpotQA dataset, RULER-HotpotQA~\cite{yu2026memagent}, by including different context lengths of articles for test questions.
The number of articles ranges from 50, 100, up to 6400, corresponding to context lengths of approximately 7K, 14K, and up to 3.5M tokens, respectively.
In the pilot study on this dataset, the length of dynamic memory cache is set to 1,024 tokens, while the chunk length is configured to 5,000 tokens.

\noindent\textbf{InstructRec-Amazon}.
Our datasets are sourced from the Amazon Review Data\footnote{\url{https://amazon-reviews-2023.github.io/index.html}} repository.
We filter the datasets using the standard 5-core criterion, removing users and items with fewer than five associated interactions to ensure sufficient data density.
We then retain only positive interactions by discarding items with review ratings below 4. Furthermore, duplicate interactions are removed.
Following prior recommendation studies~\cite{kang2018self}, we adopt the leave-one-out protocol for dataset splitting, using all but the last interaction in each user's interaction history for training, while reserving the last interaction for evaluation.
The generation of instructions and user intents follows the procedures proposed in InstructRec~\cite{zhang2026recommendation} and iAgent~\cite{xu2025iagent}.

\subsection{Baselines}
\label{app:baseline}
In the section, we provide simple introductions to the compared models in our experiments.

\begin{itemize}[leftmargin=*] 

\item \textbf{SASRec}~\cite{kang2018self} is a self-attention-based sequential recommendation model. For the \emph{Judging} task, we first compute a score for each candidate by taking the dot product between the embedding predicted by SASRec and the candidate item embedding. If the score is greater than 0.3, the prediction is considered "Yes"; otherwise, "No".

\item \textbf{BERT4Rec}~\cite{sun2019bert4rec} is a bidirectional Transformer-based sequential recommendation model trained with the BERT-style cloze objective. Both SASRec and BERT4Rec use only item IDs as input, without textual descriptions or item images. For the \emph{Judging} task, BERT4Rec uses the same decision rule as SASRec.

\item \textbf{P5}~\cite{geng2022recommendation} is a pioneering work on LLM-based RecSys, which describes recommendation tasks in a text-to-text format and employs LLMs to capture deeper semantics for personalization and recommendation. In our experiments, we deploy sequential indexing (SID) on the P5 model.

\item \textbf{TokenRec}~\cite{qu2025tokenrec} is a recently-developed for large language model-based recommendation model, which tokenizes item-side information, e.g., collaborative embeddings or textual descriptions, into several discrete tokens via VQ-VAE.
To adapt TokenRec to the \emph{Judging} task, we train it from scratch as a variant to produce a binary output ("Yes" or "No").

\item \textbf{ToolRec}~\cite{zhao2024let} incorporates attribute-oriented tools and a memory strategy into large language model-based recommender systems. It employs LLMs to closely model user preferences, thereby improving the accuracy of recommendations generated during user decision simulation.

\item \textbf{iAgent}~\cite{xu2025iagent} is an instruction-aware recommendation agent capable of using tools to simulate user behaviors and acquire knowledge from external environments.

\item \textbf{QwenLong-L1}~\cite{wan2025qwenlong} is the first long-context large reasoning model (LRM) trained with reinforcement learning for long-context reasoning. In its standard configuration, QwenLong-L1-32B supports context lengths of up to 32,768 tokens. We prompt it to solve our recommendation tasks in a zero-shot setting.

\item \textbf{Mem0}~\cite{chhikara2025mem0} is a representative \emph{Long-Term Memory} architecture that dynamically extracts, consolidates, and retrieves salient information from ongoing conversations. Mem0 encodes memory chunks into text embeddings using \emph{gpt-4o-mini} and \emph{text-embedding-3-small}.

\item \textbf{MemAgent}~\cite{yu2026memagent} targets general long-context document understanding by processing documents in chunks and iteratively updating a fixed-length memory.
\end{itemize}

\subsection{OCR Evaluation}
\label{app:ocr_eval}
\begin{figure*}[h]
\centering
{\subfigure[Games, Example 1, Item Title: \{\emph{Sword Art Online: Lost Song - PlayStation 4}\}]
{\includegraphics[width=0.45\linewidth]{{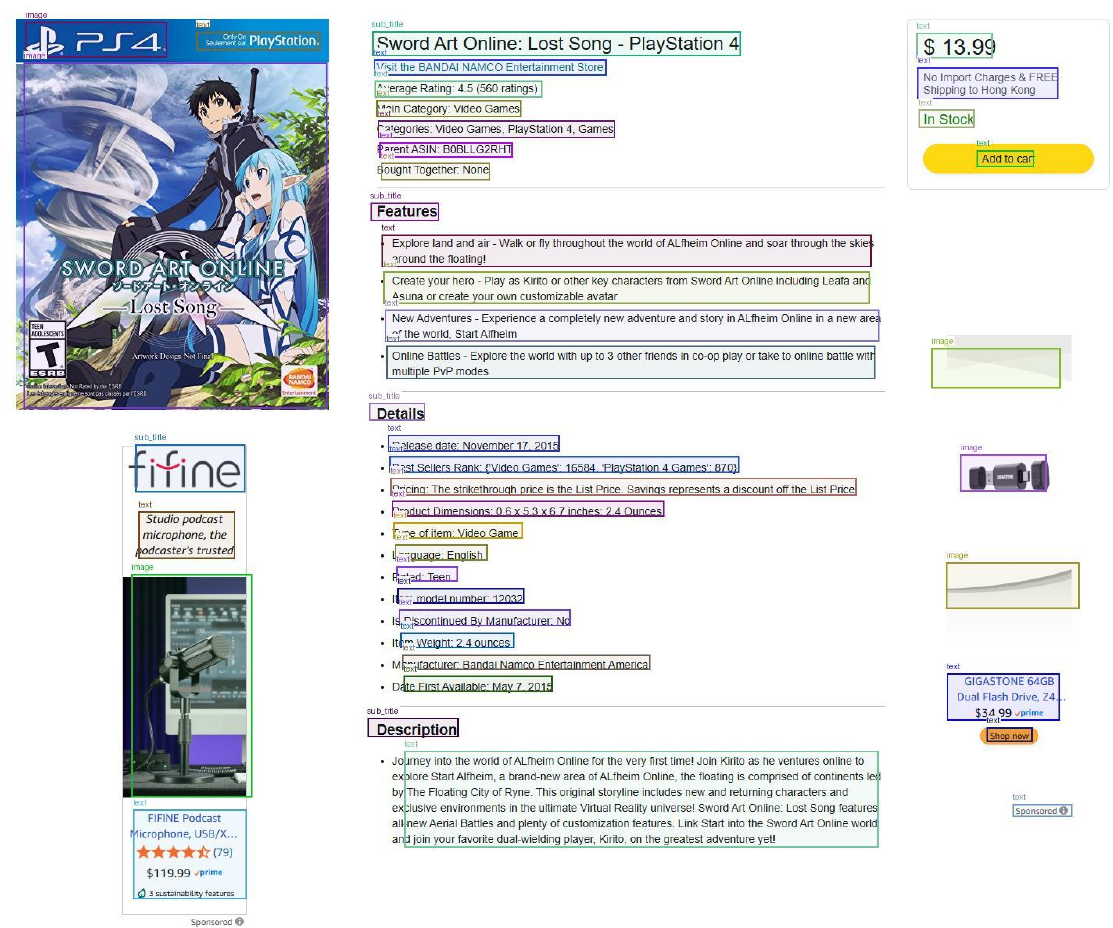}}}}
\quad
{\subfigure[Games, Example 2, Item Title: \{\emph{Smatree Charging Dock}\}]
{\includegraphics[width=0.45\linewidth]{{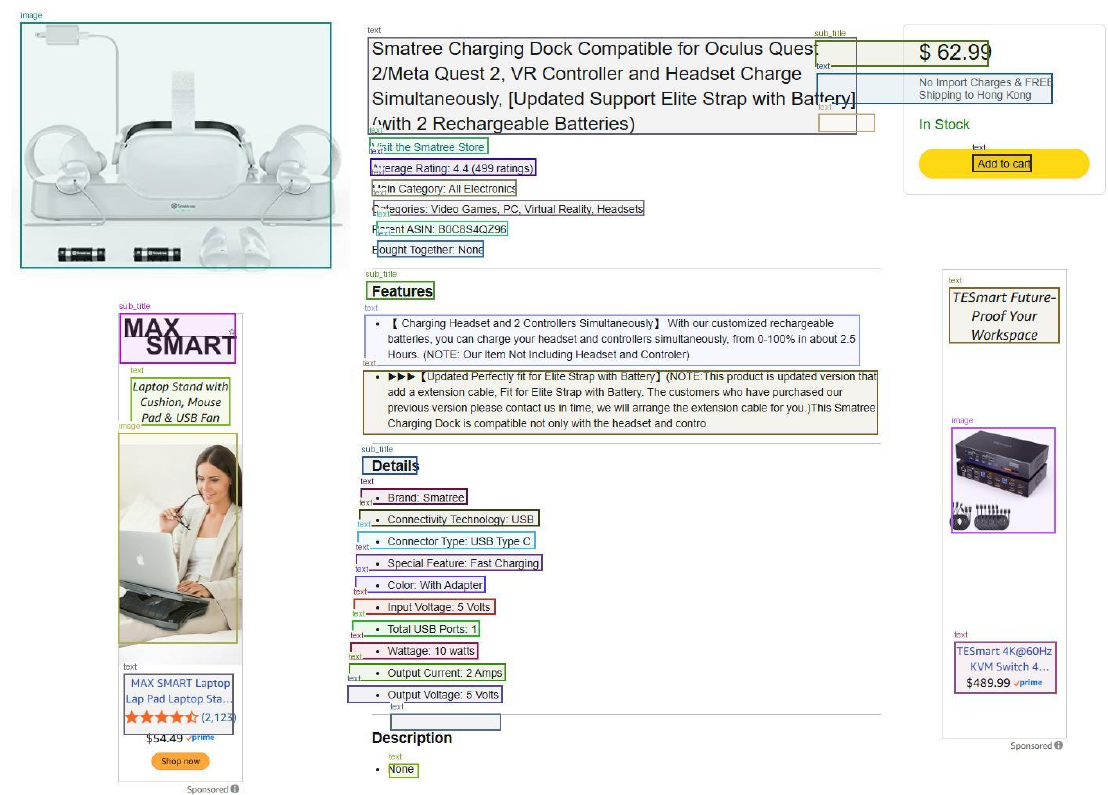}}}}
\quad
{\subfigure[WebWalkerQA, Example 1, Question: \{\emph{Which committee addresses the transition from relief to development at the NMUN Washington D.C. conference, and what is the payment deadline?}\}]
{\includegraphics[width=0.45\linewidth]{{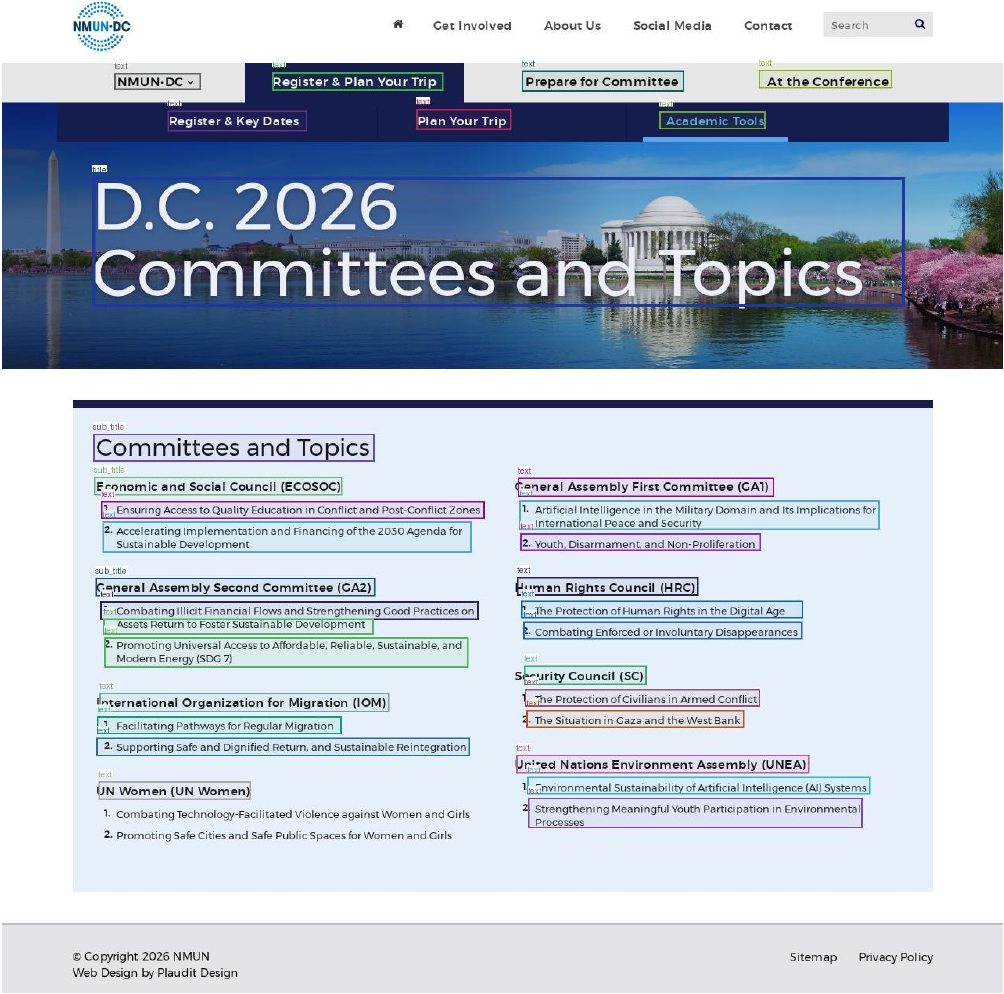}}}}
\quad
{\subfigure[WebWalkerQA, Example 2, Question: \{\emph{By when should an international attendee applying for a B1 visa in May 2024 receive their visa to attend ACR Convergence 2024?}\}]
{\includegraphics[width=0.45\linewidth]{{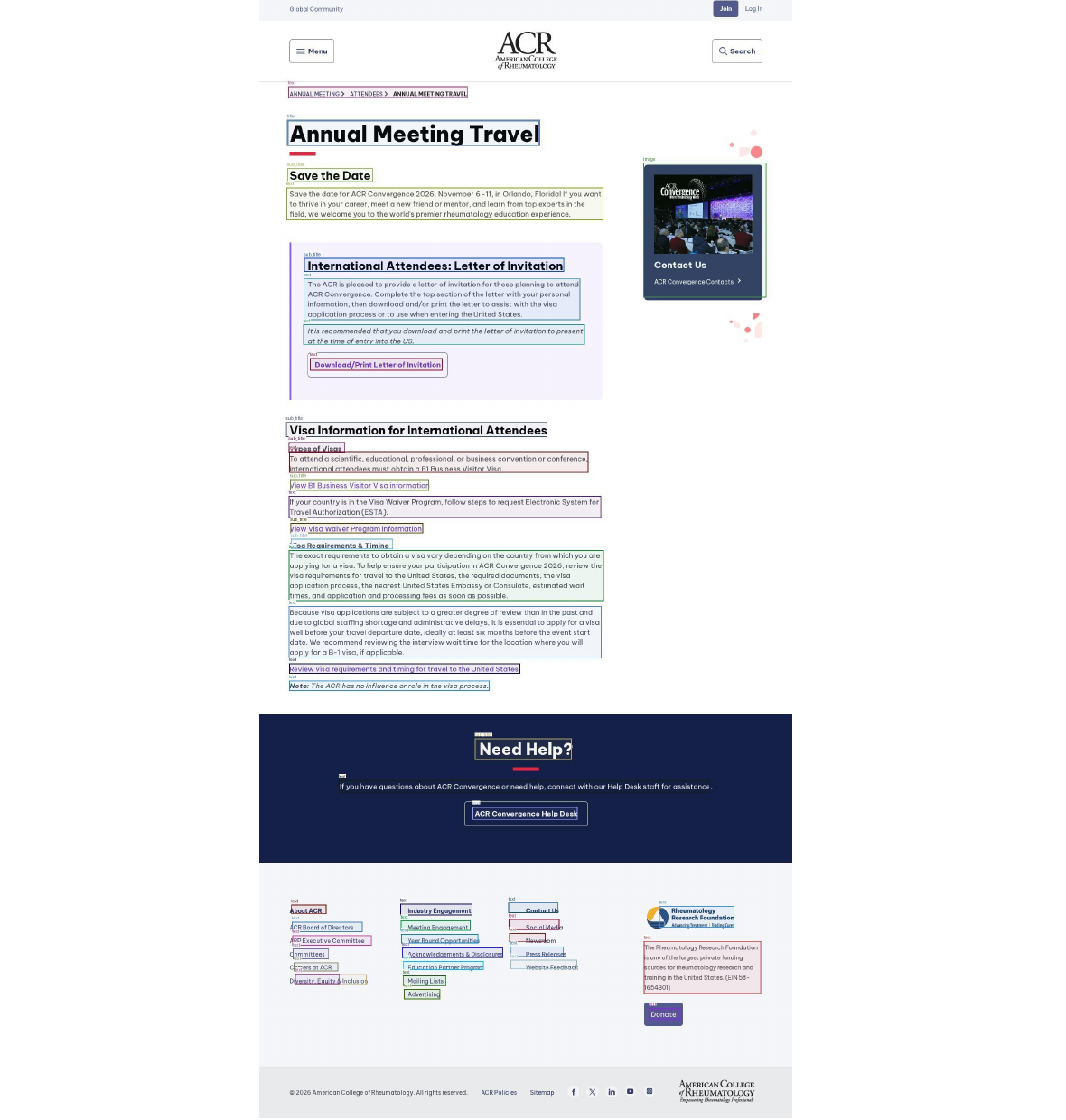}}}}
\captionsetup{font={small}}
\caption{Item-page and WebWalkerQA examples with DeepSeek-OCR-2 perception outputs~\cite{wei2026deepseek}.}\label{fig:ocr_examples}
\end{figure*}

We evaluate OCR on three recommendation datasets and WebWalkerQA (Web) using webpage snapshots; DeepSeek-OCR-2 produces Markdown descriptions and cropped page images (Figure~\ref{fig:ocr_examples}).
The OCR output consists of a Markdown (.mmd) file describing the webpage content, together with cropped images extracted from the page.
This structured information is provided to RecAgents to support preference modeling and recommendation reasoning.
Item pages are rendered using a shared shopping-page template with randomly inserted advertisements, allowing fair and controlled comparison of perception methods while limiting conclusions about naturally heterogeneous interfaces.
Table~\ref{tab:ocr_comparison} shows that both OCR variants consistently outperform plain-text input and vision-language input, with DeepSeek-OCR-2 performing best on every metric.
Removing noise also improves the text baseline, confirming that irrelevant page content impairs downstream reasoning.
Further model-specific analysis appears in the DeepSeek-OCR-2 report~\cite{wei2026deepseek}.

\begin{table*}[h]
    \centering
    \captionsetup{font={small}}
    \caption{Comparison of perception modules. The best and second-best results are highlighted in bold and underlined, respectively.}
    \label{tab:ocr_comparison}
    \small
    \setlength{\tabcolsep}{5pt}
    \scalebox{0.85}{
    \begin{tabular}{lccccccc}
        \toprule
        \multirow{2}{*}{\textbf{Module}}
        & \multicolumn{2}{c}{\textbf{Games}}
        & \multicolumn{2}{c}{\textbf{MovieTV}}
        & \multicolumn{2}{c}{\textbf{Books}}
        & \multicolumn{1}{c}{\textbf{Web}} \\
        & \textbf{HR@1} & \textbf{HR@3}
        & \textbf{HR@1} & \textbf{HR@3}
        & \textbf{HR@1} & \textbf{HR@3}
        & \textbf{Acc. (\%)} \\
        \midrule
        DeepSeek-OCR-2
        & \textbf{0.4449} & \textbf{0.5753}
        & \textbf{0.2610} & \textbf{0.3740}
        & \textbf{0.3663} & \textbf{0.5133}
        & \textbf{60.25} \\

        PaddleOCR
        & \underline{0.4202} & \underline{0.5416}
        & \underline{0.2518} & \underline{0.3540}
        & \underline{0.3502} & \underline{0.4905}
        & \underline{59.36} \\

        Vision (Qwen3-VL-8B)
        & 0.4108
        & 0.5231
        & 0.2543
        & 0.3633
        & 0.3334
        & 0.4819 
        & 55.77 \\
        
        Text (noise removed)
        & 0.3983 & 0.5175
        & 0.2506 & 0.3494
        & 0.3241 & 0.4664
        & 54.91 \\
        
        Text (with noise)
        & 0.3959 & 0.5067
        & 0.2466 & 0.3403
        & 0.3145 & 0.4520
        & 50.94 \\
        \bottomrule
    \end{tabular}
    }
\end{table*}

\subsection{Case Study}
\label{app:case}

Table~\ref{tab:case} illustrates how memory evolves from a broad interest in Xbox and VR products to a more recent preference for portable VR power accessories.
By retaining this shift, \ourname{} correctly selects \emph{VINDIJA Head Strap for Oculus Quest 2 with Battery}, showing that the memory preserves stable interests while incorporating recent behavior.

\begin{table*}[htbp]\scriptsize
  \centering
  \captionsetup{font={small}}
  \caption{A case study demonstrating the effectiveness of the proposed memory mechanism in modeling evolving user preferences.
  To improve readability, only the item titles are shown in the interaction sequence.}
  \scalebox{1}{
    \begin{tabularx}{\textwidth}{>{\hsize=0.5\hsize}C|>{\hsize=0.5\hsize}C}
    \toprule
    \multicolumn{2}{c}{\textbf{Context Length = 37,781 Tokens}} \\
    \textbf{Item Sequence $\downarrow$} & \textbf{Memory} \\
    \midrule
    0. Forza Horizon 3 \textbackslash{}u2013 Xbox One 
    
    1. RIG 400HX Stereo Gaming Headset for Xbox One

    2. Silicone Cover Set for Oculus Quest 2
    
    & $\mathbf{m}^1 =$ {The user interacts with gaming-related items, including Xbox One games (Forza Horizon 3), Xbox accessories (RIG 400HX Stereo Gaming Headset), and VR accessories (Silicone Cover Set for Oculus Quest 2). They show interest in gaming hardware and peripherals, particularly for Xbox and VR platforms. Reviews indicate they prioritize functionality and comfort (e.g., headset fit, VR cover comfort), and are open to budget-friendly options. They may also have family-oriented gaming interests.} \\
    3. USB C Portable Charger 5000mAh

    4. Oculus Quest 2 Battery Pack
    & $\mathbf{m}^2 =$ {\{Similar Content\}. Additionally, they value portable and convenient accessories that enhance gaming experiences, such as \textbf{battery packs for VR headsets (Oculus Quest 2)} to extend playtime and avoid cable clutter, and compact power banks for mobile devices during gaming sessions.} \\
    \midrule
    \multicolumn{2}{c}{\textbf{Candidates}} \\
    \multicolumn{2}{c}{\underline{VINDIJA Head Strap for Oculus Quest 2 with Battery}} \\
    \multicolumn{2}{c}{Game racing wheel 270 degree} \\
    \multicolumn{2}{c}{Pillars of Eternity II: Deadfire - Xbox One} \\
    \multicolumn{2}{c}{Smatree Carrying Case for Super NES Classic/SNES Classic Mini (2018)} \\
    \multicolumn{2}{c}{HDE 2 Pack of 128 MB Gaming Memory Cards for Nintendo Wii and Gamecube (Black)} \\
    \multicolumn{2}{c}{iMP Tech Trigger Treadz (PS4)} \\
    \multicolumn{2}{c}{Trine} \\
    \multicolumn{2}{c}{The Legend of Zelda: Twilight Princess} \\
    \multicolumn{2}{c}{Lords of the Fallen - PlayStation 4} \\
    \multicolumn{2}{c}{Giana Sisters Twisted Dreams Owltimate Edition NSW - Nintendo Switch} \\
    \midrule
    \multicolumn{2}{c}{\textbf{Answer}} \\
    \multicolumn{2}{c}{VINDIJA Head Strap for Oculus Quest 2 with Battery (Correct)} \\
    \bottomrule
    \end{tabularx}%
    }
  \label{tab:case}%
\end{table*}%

\subsection{Supplements on Various Length Reasoning}
\label{app:VarLenEval}

Table~\ref{tab:varleneval_app} presents a fine-grained evaluation of our framework across reasoning trajectories of different lengths. Overall, our method consistently outperforms the standard QwenLong-L1-32B (QwenL1) baseline on all three tasks and across both datasets, demonstrating its robustness to increasingly complex reasoning processes.
As expected, performance gradually declines as the reasoning length increases from Short to Long.
This trend is particularly evident on the Games dataset, where the number of reasoning tokens expands from 0--14K to over 112K.
Longer trajectories introduce substantially more contextual information, increasing the difficulty of identifying relevant evidence and maintaining coherent reasoning over extended contexts. Nevertheless, our method preserves a clear advantage over QwenL1 in all settings. 
For example, on the searching task, our approach improves HR@1 from 0.3005 to 0.3563 (+18.6\%) on the long Games subset, while on the ranking task it raises NDCG@3 from 0.3449 to 0.4624 (+34.1\%).
Similar gains are observed for the judging task, where the accuracy advantage remains around four percentage points even under the longest reasoning contexts.
A similar trend is observed on the MovieTV dataset despite its considerably smaller number of long-context samples (only 47 instances). Although performance also decreases with reasoning length, our method consistently achieves higher HR@1, NDCG@3, and judgment accuracy than the baseline. Notably, the relative improvements on the long subset remain substantial, indicating that the proposed framework generalizes well across domains and is less susceptible to performance degradation caused by long reasoning chains.
These results suggest that the proposed reasoning framework is more effective at filtering irrelevant information, preserving long-range dependencies, and leveraging extensive contextual evidence than directly applying the base language model.
Consequently, it exhibits stronger scalability to complex recommendation scenarios that require prolonged reasoning and multi-step decision making.

\begin{table*}[t]
\centering
  \captionsetup{font={small}}
    \caption{Supplements on various length reasoning evaluation across three recommendation tasks on the Games and MovieTV datasets.}
    \scalebox{0.85}{
    \begin{tabular}{c|ccc|ccc}
        \toprule
       \multirow{2}{*}{\textbf{Dataset}} & \multicolumn{3}{c}{\textbf{Games}} & \multicolumn{3}{c}{\textbf{MovieTV}} \\ \cmidrule(lr){2-4} \cmidrule(lr){5-7}
        & \textbf{Short}      & \textbf{Medium}      & \textbf{Long}      & \textbf{Short}      & \textbf{Medium}      & \textbf{Long}      \\ \midrule
        \# Samples &  25,550 & 68,350 & 862 & 482 & 5,120 & 47  \\
        \# Tokens (K) &  0-14 & 14-112 & 112-722 & 0-14 & 14-112 & 112-292  \\ \midrule
        \multicolumn{7}{c}{\textbf{Task: Searching, Metric: HR@1}} \\
        \midrule
        Ours & 0.4462 & 0.4372 & 0.3610 & 0.2665 & 0.2605 & 0.2261  \\
        QwenLong-L1-32B & 0.3418 & 0.3113 & 0.3005  & 0.2363 & 0.2227 & 0.2072  \\ \midrule
\multicolumn{7}{c}{\textbf{Task: Ranking, Metric: NDCG@3}} \\
        \midrule
        Ours & 0.5472 & 0.5231 & 0.4729 & 0.3933 & 0.3877 & 0.3342 \\
        QwenLong-L1-32B &  0.4531 & 0.4388 & 0.3449  & 0.3461 & 0.3344 & 0.2971 \\
\midrule
\multicolumn{7}{c}{\textbf{Task: Judging, Metric: Accuracy (\%)}} \\
        \midrule
        Ours & 0.8455 & 0.8331 & 0.7558 & 0.7429 & 0.7347 & 0.6853  \\
        QwenLong-L1-32B &   0.8067 & 0.7928 & 0.7143  & 0.7053 & 0.6795 & 0.6109 \\
        \bottomrule
    \end{tabular}
    }
    \label{tab:varleneval_app}
\end{table*}

\subsection{Hyperparameter Analysis}
\label{app:param}

\begin{figure*}[t]
\centering
{\subfigure[Searching]
{\includegraphics[width=0.31\linewidth]{{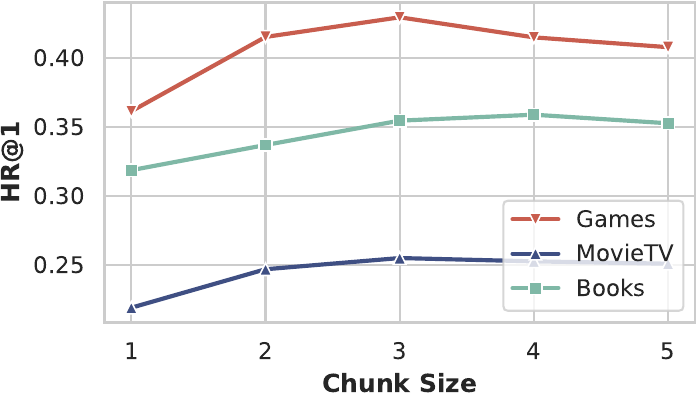}}}}
\quad
{\subfigure[Ranking]
{\includegraphics[width=0.31\linewidth]{{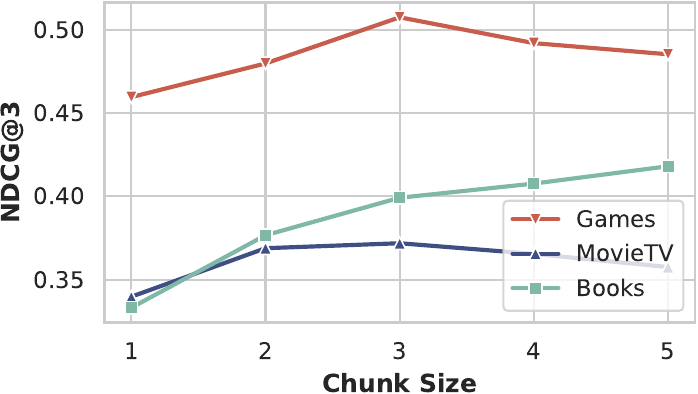}}}}
\quad
{\subfigure[Judging]
{\includegraphics[width=0.31\linewidth]{{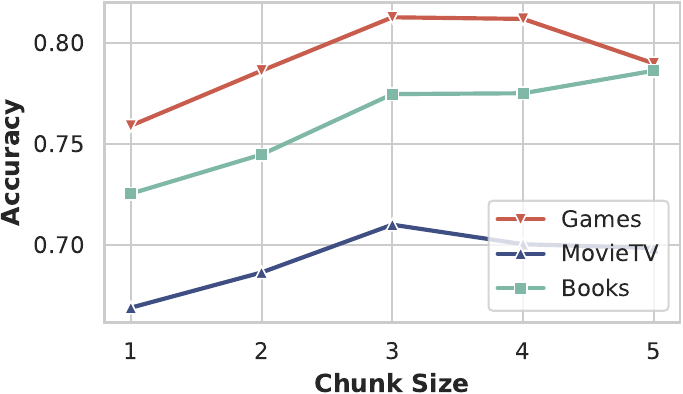}}}}
\vskip -0.1in
\captionsetup{font={small}}
\caption{Effect of chunk window size $W$.}\label{fig:chunk_size}
\end{figure*}

Figure~\ref{fig:chunk_size} shows that performance generally improves up to $W=3$ and then plateaus or declines.
$W=3$ is best or near-best for Games and MovieTV, whereas the longer Books histories favor $W=4$ or $5$, indicating that moderately larger chunks benefit longer sequences.

\subsection{Inference Latency}
\label{app:latency}

Table~\ref{tab:latency} reports two operating regimes. In the intended deployment setting, \ourname{} maintains and incrementally updates a persistent user memory; \ourname{}-Cache therefore measures online recommendation latency using the cached memory $\mathbf{m}^{T}$. It requires $1.44$--$3.74$ seconds per request while keeping the active context bounded. In contrast, \ourname{}-Build reconstructs the memory from the complete supplied history before generating an answer. We include this setting as a reference for the full processing workflow rather than as the expected per-request serving mode.

The cached and full-history rows cover different computational scopes and should not be interpreted as a direct end-to-end speed comparison. Moreover, the current logs do not separately record OCR preprocessing or incremental memory-update latency. Thus, the results demonstrate bounded per-request context and report final-answer latency in the intended cached-memory regime, but do not establish an amortized or end-to-end efficiency advantage.

\begin{table*}[h]
  \centering
  \captionsetup{font={small}}
  \caption{Average latency (seconds/sample) on one H20 GPU. \ourname{}-Build builds memory; \ourname{}-Cache assumes cached memory.}
  \scalebox{0.85}{
    \begin{tabular}{ccccc}
    \toprule
     & \textbf{History state} & {\textbf{Books}} & {\textbf{MovieTV}} & {\textbf{Games}} \\
   \textbf{Method} & (Avg. input tokens) & (62,281.48) & (30,207.68) & (26,217.33) \\
    \midrule
    Qwen3.5-9B & Supplied history & 9.81 & 3.95 & 3.63 \\
    QwenLong-L1-32B & Supplied history & 108.91 & 47.99 & 42.41 \\
    \midrule
    \ourname{}-Build & Build from history & 224.60 & 118.53 & 98.25 \\
    \ourname{}-Cache & Cached $\mathbf{m}^T$ & 3.74 & 1.55 & 1.44  \\
    \bottomrule
    \end{tabular}%
    }
  \label{tab:latency}%
\end{table*}%

\section{Discussion on Limitations and Future Work}
\label{app:future_work}
This work focuses on two components of the recommendation-agent pipeline: extracting item evidence from rendered pages and modeling user preferences from long interaction histories.
It does not evaluate full-catalog retrieval, live website navigation, or closed-loop interaction with users.
Claims concerning arbitrary websites, cross-platform deployment, and closed-loop user interaction remain outside the scope of the present evaluation.
Looking forward, we plan to extend the current recommendation environment with richer user-agent interactions.
While this work primarily models common user behaviors on item pages, such as \emph{browsing product images}, \emph{comparing multiple items}, and \emph{making purchasing decisions}, real-world recommendation systems involve a much broader spectrum of user actions, including expanding item descriptions, searching for additional information, and completing the checkout process.
Incorporating these fine-grained behaviors into the agent’s action space would enable more realistic user simulations and provide richer signals for preference modeling.

\end{document}